\documentclass[10pt,twocolumn,letterpaper]{article}

\usepackage[pagenumbers]{cvpr} 

\usepackage[utf8]{inputenc} 
\usepackage[T1]{fontenc}    
\usepackage{url}            
\usepackage{amsfonts}       
\usepackage{nicefrac}       
\usepackage{microtype}      
\usepackage{graphicx}
\usepackage{amsmath}
\usepackage{amssymb}
\usepackage{booktabs}
\usepackage{tabularx}
\usepackage{capt-of}
\usepackage{hyperref}  
\usepackage[dvipsnames]{xcolor}

\title{Reconstruction of a 3D wireframe from a single line drawing \\ via generative depth estimation}

\author{
    Elton Cao \qquad Hod Lipson \\[1.5ex]
    Creative Machines Lab, Columbia University \\
    New York, NY \\
}

\begin{document}

\twocolumn[{%
\renewcommand\twocolumn[1][]{#1}%
\maketitle

\begin{center}
    \includegraphics[width=0.95\textwidth]{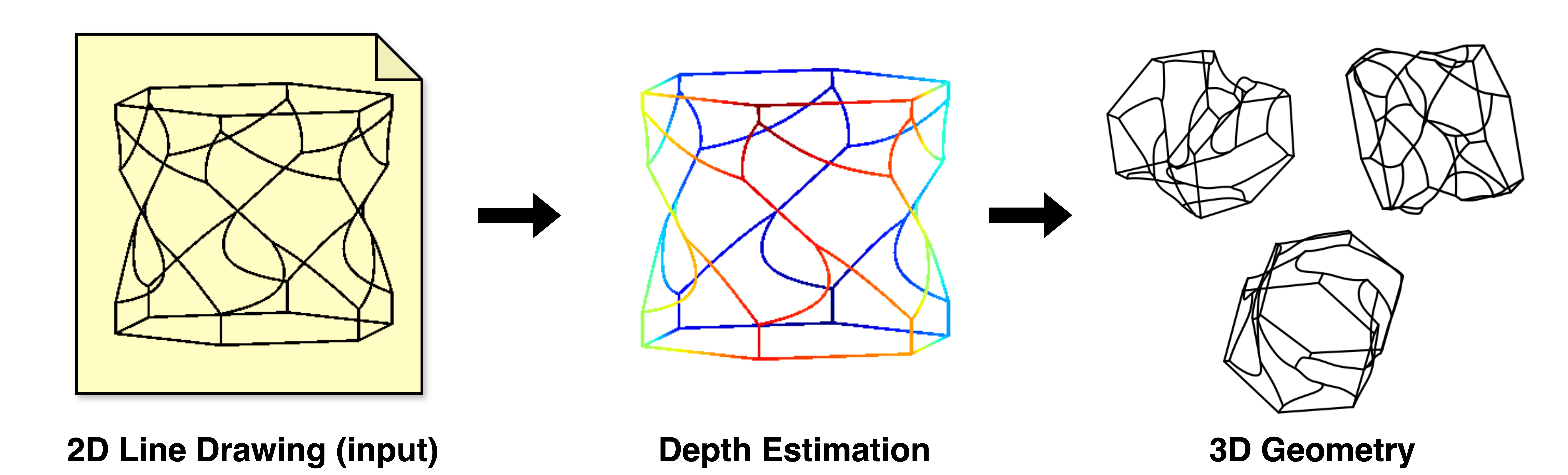} 
    \captionof{figure}{\textbf{The engineering line drawing reconstruction challenge.} A single orthographic line drawing presents numerous ``possible'' reconstructions, some more plausible than others. Our method predicts depth only on the drawn pixels, producing a projection-faithful 3D wireframe rather than completing unseen geometry or inferring a full CAD program.}
    \label{fig:overview}
    \vspace{0.3cm} 
\end{center}%
    }]

\begin{abstract}

Reconstructing 3D geometry from 2D engineering line drawings is an inherently ambiguous problem: while visible strokes determine the object’s projected structure, they do not specify the depth of each stroke. Rather than treating this problem as sketch-based asset generation, where models often infer unobserved structure, we study projection-faithful wireframe reconstruction: lifting a user-provided drawing into 3D according to its visible strokes. We formulate this task as conditional depth estimation over line drawings, predicting a depth value for each drawn pixel to produce a 3D wireframe. To model the ambiguities of orthographic projection, we implement a Latent Diffusion Model with spatial conditioning on the input sketch and optional partial-depth conditioning for iterative reconstruction. We train and evaluate our models on over three million synthetic image-depth pairs derived from CAD wireframes, including a newly curated corpus of roughly 90,000 shapes. Across varying shape complexities, our framework achieves robust reconstruction performance; scaling from $256$ to $512$ resolution with a retrained latent space roughly halves reconstruction error, reaching a 3.9\% best-of-five (7.0\% average) normalized depth error. These results demonstrate the potential of projection-faithful depth estimation as a user-controlled approach for iterative 3D wireframe creation in engineering design.

\end{abstract}

\section{Introduction}

Sketching is a natural and efficient medium for communicating geometric ideas, especially during the early stages of engineering design~\cite{eissen2008, olsen2009, tono2026}. With a small number of strokes, a designer can express object structure, spatial relationships, and intended form before the object is fully specified in a formal modeling system~\cite{sutherland1963, olsen2009}. Converting such 2D drawings into 3D geometry, however, remains a long-standing challenge: a line drawing constrains the projected structure of an object, but generally does not uniquely determine the depth, topology, or spatial ordering of its strokes~\cite{huffman1971, sugihara1986, lipson1996, mitra2013}. This ambiguity is especially pronounced for engineering line drawings, which are sparse, unshaded, and dominated by visible edges, contours, and wireframe-like structures rather than the photometric cues available in natural images~\cite{ranftl2020, ranftl2021, yang2024, koley2025}.

Sketch-based interfaces for modeling (SBIM) have addressed this problem through a wide range of interaction and reconstruction paradigms, including gesture-based primitive creation, contour inflation, extrusion, surface sketching, and multi-view modeling~\cite{zeleznik1996, igarashi1999, karpenko2006, nealen2007, olsen2009}. More recently, deep sketch-based 3D modeling (DS-3DM) has expanded this design space with neural networks, transformers, implicit representations, differentiable rendering, diffusion models, and foundation-model optimization~\cite{delanoy2017, zhong2020deepsketch, li2020, wu2021, li2022, sketch2mesh2021, lasdiffusion2023, control3d2023, sketchdream2024, sketch3d2024, tono2026}. These methods differ substantially in their input assumptions, viewpoint constraints, sketch styles, output representations, and evaluation metrics~\cite{tono2026}. As a result, ``sketch-to-3D'' has emerged as not a single benchmark problem, but a family of related tasks with distinct output contracts.

We use output contract to refer to the representation a method is designed to produce, such as CAD command sequences, face graphs, watertight meshes, implicit fields, point clouds, textured assets, or depth values over visible sketch support. This distinction is central to engineering line drawing reconstruction. Many recent sketch-conditioned 3D methods use a sketch as a prompt for generating a complete semantic object, often inferring unseen structure, parts, textures, or appearance~\cite{sketch2mesh2021, sketchsampler2022, lasdiffusion2023, control3d2023, sketchdream2024, sketch2nerf2024, sketch3d2024, tono2026}. Such methods are powerful for asset generation and conceptual modeling. In contrast, engineering drawing reconstruction often requires a projection-faithful representation: the system should preserve the geometry explicitly drawn by the user~\cite{lipson1996, li2020, li2022, wang2022}.

In this paper, we study projection-faithful 3D reconstruction of engineering line drawings. Given a sparse 2D drawing of a designed object, our goal is to preserve the visible stroke support and infer depth values that embed the drawing as a 3D wireframe, allowing a user to "draw in 3D". Thus, the resulting model predicts depth for drawn pixels rather than producing a CAD program, face graph, or completed semantic asset. This formulation provides a direct route from 2D to 3D for engineering line drawings while avoiding assumptions that the input must correspond to a complete CAD command sequence, a watertight surface, or a valid set of closed faces.

Alternatively, existing engineering-sketch reconstruction systems impose more restrictive assumptions. CAD-command approaches such as Sketch2CAD, DeepCAD, and Free2CAD represent geometry as sequences of parametric operations, stroke groupings, or sketch-extrude programs, enabling editability and structure but constraining the output to a predefined vocabulary~\cite{li2020, wu2021, li2022}. Topological methods such as Neural Face Identification reconstruct face connectivity from wireframe projections, but require the drawing to support valid face loops or manifold-style structure~\cite{wang2022}.

We formulate engineering line drawing reconstruction as conditional depth estimation over the visible sketch support. If a depth value is predicted for each drawn pixel, the 2D drawing can be lifted directly into an embedded 3D wireframe while preserving the input projection. This image-based formulation aligns with SBIM systems that represent strokes as raster images for compatibility with image-processing pipelines~\cite{olsen2009}, and it avoids requiring vectorization, stroke grouping, primitive fitting, or face identification as a prerequisite for reconstruction~\cite{lipson1996, li2020, li2022, wang2022}. Because orthographic line drawings lack shading, texture, perspective, and natural-image context, multiple 3D depth assignments may remain plausible for the same 2D support~\cite{huffman1971, sugihara1986, olsen2009, mitra2013}. We therefore model the task generatively using a Latent Diffusion Model (LDM) conditioned on the input drawing and, when available, partial-depth information from previous reconstruction steps~\cite{ho2020, rombach2022, ke2024}.

The partial-depth conditioning supports an iterative sketch-reconstruct-sketch workflow. A user can draw an initial set of strokes, reconstruct them in 3D, and then add new strokes conditioned on the existing 3D structure. This follows the progressive modeling goals of SBIM while using depth-space geometric context rather than CAD-command context~\cite{olsen2009, li2020, li2022}. Thus, the result is a projection-faithful mechanism for lifting sparse engineering line drawings into 3D without requiring a full CAD kernel, feature-history tree, or semantic 3D generator.

Our contributions are as follows:
\begin{itemize}
\item A projection-faithful formulation of engineering line-drawing reconstruction as conditional depth estimation on visible sketch support.
\item A diffusion-based reconstruction model with spatial sketch conditioning and partial-depth conditioning for iterative sketching.
\item Two large-scale rendered datasets derived from ABC geometry: over one million wireframe/depth pairs from a filtered prior-work subset, and a newly curated curve-rich corpus of roughly 90,000 shapes yielding 2.1 million pairs.
\end{itemize}

\section{Related Work}

\begin{table*}[t!]
\centering
\sffamily
\footnotesize
\renewcommand{\arraystretch}{1.1}
\caption{\textbf{Most closely related work for engineering line drawing reconstruction.}
We compare methods that reconstruct 3D structure from line drawings, wireframes, or CAD-style sketches of designed objects. Because these methods produce different output representations, direct metric-to-metric comparison is difficult. The table emphasizes each method's output contract and assumptions. The ``Shape'' column shows a reconstruction among the most complex demonstrated in each publication (images reproduced from the cited papers); for ours, the cell shows an actual reconstruction by our model of a 60-stroke test sketch in which 34 strokes are free-form B-spline curves.}
\label{tab:related_works}
\begin{tabularx}{\textwidth}{p{2.05cm} p{1.95cm} X X p{1.80cm} p{1.55cm} p{0.80cm}}
\toprule
\textbf{Reference}
& \textbf{Input domain}
& \textbf{Technique}
& \textbf{Key assumptions / limitations}
& \textbf{Output contract}
& \textbf{Shape}
& \textbf{Code} \\
\midrule

Huffman (1971)~\cite{huffman1971}
& Polyhedral line drawings
& Qualitative line labeling for concave, convex, and occluding edges.
& Determines symbolic consistency, but does not recover a 3D shape.
& Line labels
& ---
& N/A \\

Sugihara (1986)~\cite{sugihara1986}
& Polyhedral line drawings
& Geometric analysis of whether a line drawing can correspond to a valid 3D object.
& Studies constraints, but does not provide a learned reconstruction pipeline.
& Geometric feasibility
& ---
& N/A \\

Lipson and Shpitalni (1996)~\cite{lipson1996, lipson2007}
& Engineering line drawings / wireframes
& Optimizes vertex depths using geometric regularities such as planarity, parallelism, orthogonality, and symmetry.
& Requires clean vector graphs and reliable vertex-edge structure, but brittle for ambiguous sketches.
& 3D wireframe graph
& \raisebox{-0.85\height}{\includegraphics[width=1.45cm]{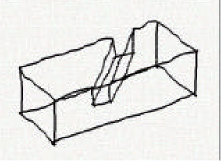}}
& N/A \\

Sketch2CAD (2020)~\cite{li2020}
& CAD-style sketches
& Parses sketches into sequential CAD modeling operations and parameters.
& Produces editable CAD-like programs, but is constrained by a fixed vocabulary.
& CAD command sequence
& \raisebox{-0.85\height}{\includegraphics[width=1.45cm]{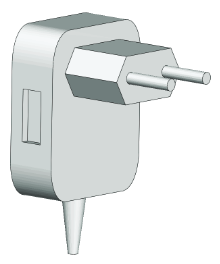}}
& Open \\

DeepCAD (2021)~\cite{wu2021}
& CAD construction sequences
& Learns a generative model over CAD command sequences.
& Models CAD programs rather than arbitrary input line drawings \& constrained by a fixed vocabulary.
& CAD command sequence
& \raisebox{-0.85\height}{\includegraphics[width=1.45cm]{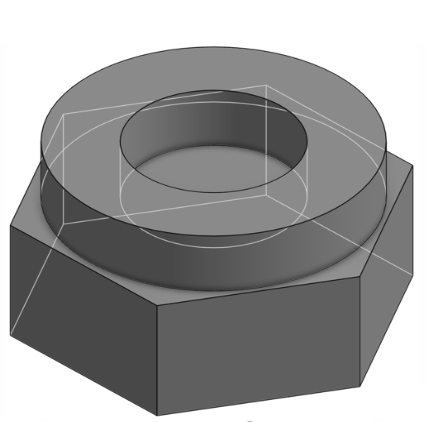}}
& Open \\

Free2CAD (2022)~\cite{li2022}
& Freehand CAD sketches
& Groups strokes into geometric compounds and fits CAD operations.
& Handles rougher sketches than command-only methods, but still constrained by a fixed vocabulary.
& CAD operations
& \raisebox{-0.85\height}{\includegraphics[width=1.45cm]{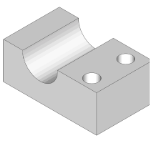}}
& Open \\

Neural Face Identification (2022)~\cite{wang2022}
& CAD-derived wireframes
& Predicts face loops and topological connectivity from 2D wireframe drawings.
& Assumes the drawing supports valid closed faces and face connectivity.
& Face graph
& \raisebox{-0.85\height}{\includegraphics[width=1.45cm]{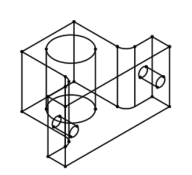}}
& Open \\

Ours
& Engineering line drawings / CAD-derived wireframes
& Predicts depth values over visible sketch support using conditional generative depth estimation.
& Does not recover a CAD command sequence, explicit face topology, or watertight surface.
& Projection-faithful 3D wireframe depth
& \raisebox{-0.85\height}{\includegraphics[width=1.45cm]{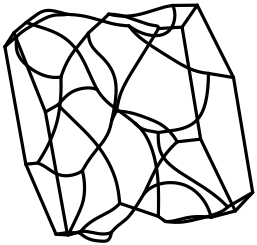}}
& Open \\

\bottomrule
\end{tabularx}
\end{table*}


\subsection{Sketch-Based Modeling and Deep Sketch-Based 3D Modeling}

Sketch-based interfaces for modeling have a long history in computer graphics, geometric modeling, and human-computer interaction~\cite{sutherland1963, zeleznik1996, igarashi1999, olsen2009}. Olsen et al. organize SBIM systems by how sketches are interpreted. This includes instructions to create new 3D models, augment existing models, or deform and manipulate geometry~\cite{olsen2009}. This highlights that sketch interpretation depends entirely on the intended output representation and interaction mode. Engineering-style drawings have historically motivated fitting, beautification, constraint inference, and geometric optimization because precision and structural integrity are central to the task~\cite{lipson1996, olsen2009, sutherland1963}.

Recent DS-3DM methods extend this literature with learning-based models and large paired datasets~\cite{tono2026}. Tono et al. categorize modern methods according to input amount, viewpoint assumptions, sketch style, model architecture, part structure, output options, geometry type, and evaluation metrics~\cite{tono2026}. Their survey includes neural models, implicit representations, transformers, diffusion models, differentiable rendering, and foundation-model optimization~\cite{delanoy2017, zhong2020deepsketch, sketch2mesh2021, lasdiffusion2023, control3d2023, sketchdream2024, sketch2nerf2024, sketch3d2024, tono2026}. This literature provides the broader context for our work, while also motivating a narrower comparison set: methods that produce textured assets, NeRFs, Gaussian splats, implicit fields, point clouds, CAD operations, and face graphs solve related but distinct output contracts.

\subsection{Classical Line-Drawing Interpretation}

3D reconstruction from line drawings was studied before modern learning-based modeling as a problem of symbolic interpretation, geometric feasibility, and constraint satisfaction~\cite{huffman1971, clowes1971, sugihara1986, lipson1996}. Huffman and Clowes introduced line-labeling frameworks for interpreting polyhedral scenes, while Sugihara studied the consistency and realizability of line drawings under geometric constraints~\cite{huffman1971, clowes1971, sugihara1986}. These works established a central challenge for line-drawing reconstruction: a 2D drawing may admit multiple valid 3D interpretations, and some drawings may not correspond to a physically realizable object at all~\cite{huffman1971, sugihara1986, olsen2009}.

Metric reconstruction methods introduced geometric priors to recover plausible 3D structure. Lipson and Shpitalni formulated line-drawing reconstruction as an optimization problem over vertex depths using regularities such as planarity, parallelism, perpendicularity, symmetry, and isometry~\cite{lipson1996, lipson2007}. Masry and Lipson later explored neural and evolutionary approaches for reconstructing 3D wireframes from 2D sketches, showing that learned priors could complement geometric regularity constraints~\cite{masry2005}. These works are historically close to ours because they address line-drawing-to-wireframe reconstruction rather than semantic asset generation, serving as an older lineage work to our work.

\subsection{Engineering Sketch and CAD Reconstruction}

A major branch of recent work reconstructs engineering drawings by mapping sketches into CAD-like or topology-aware representations. Sketch2CAD treats sketch interpretation as a sequence of CAD modeling operations, using existing partial geometry to decode later sketch commands~\cite{li2020}. DeepCAD learns a generative model over CAD command sequences, demonstrating that CAD construction histories can be represented as learnable sequential data~\cite{wu2021}. Free2CAD extends this direction to freehand CAD sketches by grouping strokes into geometric compounds and fitting parametric operations with transformer-based architectures~\cite{li2022, tono2026}. Related program- and parameter-based systems such as GeoCode, PICASSO, and SECAD-Net further demonstrate the value of symbolic or procedural representations for editable 3D design~\cite{geocode2022, picasso2023, secadnet2024}.

These methods are highly relevant because they target engineering-style inputs. Their strength is editability: by recovering CAD operations or parametric programs, they produce representations that can be manipulated in downstream modeling systems. Thus, their assumptions maintain that strength. The object must be expressible in the chosen vocabulary of primitives, operations, or sketch-extrude programs~\cite{li2020, wu2021, li2022}. Our method makes a different tradeoff by predicting depth on visible sketch support, sacrificing immediate CAD editability but reducing dependence on primitive vocabulary and other restraints.

Topological reconstruction methods provide another close comparison. Neural Face Identification predicts face loops from 2D wireframe projections of manifold objects, framing wireframe reconstruction as a sequence prediction problem over graph connectivity~\cite{wang2022}. Its output contract is face topology, whereas ours is depth over visible sketch pixels. The two formulations are complementary, since a face-topology model can recover surface structure when valid face loops are present, while depth lifting can operate on partial, open, or intermediate wireframes that may not yet define closed faces.

\subsection{Domain-Specific Engineering Sketch Reconstruction}

Several recent systems restrict the domain or drawing convention to make sketch-based reconstruction tractable. Sketch2PQ reconstructs planar quadrilateral meshes from axonometric architectural sketches by assuming roof-like PQ mesh structure and using explicit curve categories~\cite{sketch2pq2022}. Vitruvio reconstructs building geometry from single perspective sketches, adapting single-view reconstruction ideas to architecture-specific data~\cite{vitruvio2024, tono2026}. These methods show that engineering-specific assumptions can be powerful when the drawing convention and output representation are fixed.

\subsection{Sketch-Based 3D Asset Generation}

Many modern sketch-based 3D methods focus on object completion or asset generation. Sketch2Mesh, Sketch2Model, DeepSketch, LAS-Diffusion, Control3D, SketchDream, Sketch2NeRF, and Sketch3D map sketches to meshes, implicit fields, NeRFs, Gaussian splats, point clouds, or textured 3D assets~\cite{zhong2020deepsketch, sketch2model2021, sketch2mesh2021, lasdiffusion2023, control3d2023, sketchdream2024, sketch2nerf2024, sketch3d2024, tono2026}. These methods often use learned object priors, differentiable rendering, text prompts, foundation models, or diffusion losses to infer geometry and appearance not explicitly present in the input sketch~\cite{control3d2023, sketchdream2024, sketch2nerf2024, sketch3d2024}. This design point is well suited to creative asset generation, where a sketch may serve as an incomplete prompt for a full 3D object.

Our work occupies a different design point in the DS-3DM taxonomy. The goal is not to complete a semantic object, synthesize appearance, or infer hidden parts, but to preserve visible stroke support and estimate depth values for those strokes. Broad sketch-to-asset systems are therefore important context, but they are not direct baselines for our task.

\subsection{Depth Estimation and Generative Depth Priors}

Monocular depth estimation has traditionally been studied on natural images, where models can exploit shading, texture, perspective, object semantics, and large-scale image priors~\cite{eigen2014, ranftl2020, ranftl2021, yang2024}. Recent models such as MiDaS, DPT, Depth Anything, and Depth Anything V2 achieve strong generalization across photographic datasets, but their assumptions do not directly transfer to sparse engineering wireframes~\cite{ranftl2020, ranftl2021, yang2024, yang2024v2}. Engineering line drawings lack the photometric and semantic cues that make natural-image depth estimation effective.

Generative depth models provide a natural framework for ambiguous reconstruction. Diffusion models have achieved strong results in image generation and conditional synthesis by modeling multimodal output distributions rather than regressing a single deterministic target~\cite{ho2020, song2021, rombach2022}. Marigold adapts latent diffusion priors to monocular depth estimation, demonstrating that diffusion-based models can be effective for ill-posed depth prediction when trained or fine-tuned for the appropriate domain~\cite{ke2024}. Our work adopts this generative perspective for engineering line drawings.

\begin{figure*}[t]
    \centering
    \includegraphics[width=0.8\textwidth,height=0.4\textheight,keepaspectratio]{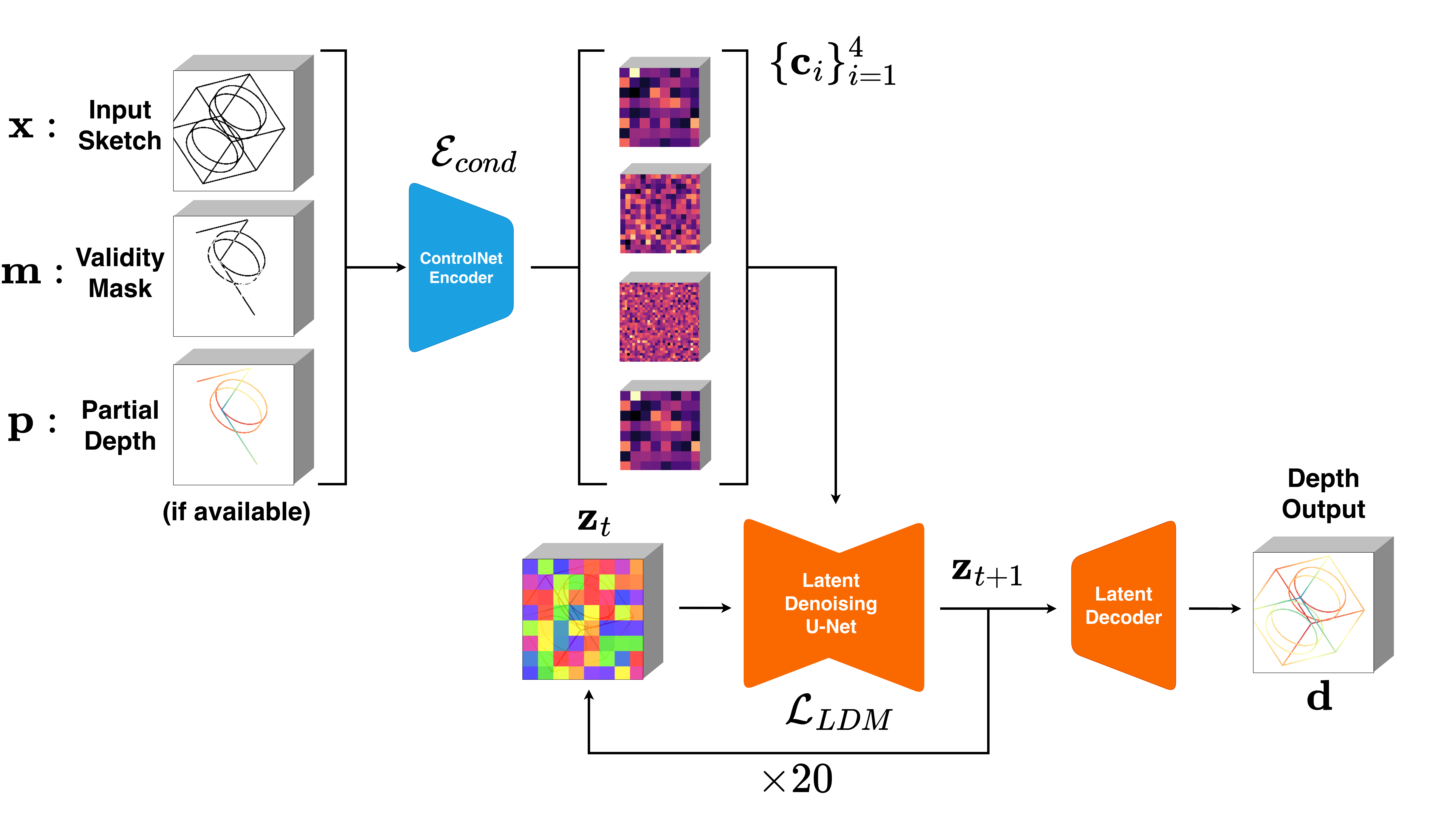}
    \caption{\textbf{LDM architecture with ControlNet conditioning.} Our model's diffusion architecture involves passing the geometric conditions ($\mathbf{x}$, $\mathbf{p}$, and $\mathbf{m}$) into our conditioning encoder, outputting representations at various resolutions, $\{\mathbf{c}_i\}_{i=1}^4$. These representations are then injected into the LDM, which predicts the Gaussian noise added to the latent space depth maps. The resulting latent space is then decoded into the depth map output.}
    \label{fig:architecture}
\end{figure*}

\section{Problem Formulation}
We frame the task of wireframe reconstruction as a conditional dense depth estimation problem. Unlike traditional CAD reconstruction, which aims to recover a sequence of parameters or a graph of discrete geometric primitives, our goal is to learn a mapping from a 2D raster sketch to an embedded 3D wireframe representation via per-pixel depth.

Let $\mathbf{x} \in \mathbb{R}^{H\times W \times C}$ represent a 2D raster sketch of a wireframe object. This input $\mathbf{x}$ is assumed to be a projection of an underlying 3D ground truth geometry $\mathcal{G}$. Our objective is to recover the depth map $\mathbf{y} \in \mathbb{R}^{H \times W}$ corresponding to the camera viewpoint of the projection $\mathcal{G}$. The value $\mathbf{y}_{u,v}$ represents the normalized distance from the camera plane to the geometry at pixel coordinates $(u,v)$.

The fundamental challenge in single-view wireframe reconstruction is that the mapping from $\mathbf{x}$ to $\mathbf{y}$ is ill-posed~\cite{huffman1971, sugihara1986, olsen2009, mitra2013}. Under orthographic projection, visual cues such as perspective foreshortening are absent. Consequently, a single 2D wireframe $\mathbf{x}$ may mathematically correspond to multiple valid 3D interpretations (e.g., the Necker Cube ambiguity). Because a deterministic regression model would tend to average these possibilities, we instead formulate the problem as learning the conditional probability distribution:
$$p(\mathbf{y} | \mathbf{x})$$

This probabilistic formulation allows the model to generate a specific and valid 3D structure by sampling from the learned distribution of valid geometries.

To support the ``sketch-reconstruct-sketch'' workflow, we introduce a secondary conditioning variable, the partial depth map $\mathbf{p}$ and a validity mask $\mathbf{m}$. These variables represent sparse or incomplete 3D information (e.g., existing geometry from a previous iteration of the design). The input to our model is thus a combination of $(\mathbf{x}, \mathbf{p}, \mathbf{m})$, where:
\begin{itemize}
    \item $\mathbf{x}$ is the current binary input sketch mask.
    \item $\mathbf{p}$ is the existing/partial depth.
    \item $\mathbf{m}$ is the validity mask indicating which partial depth pixels can be fully trusted.
\end{itemize}

The full objective is to model the conditional distribution $p(\mathbf{y} | \mathbf{x}, \mathbf{p}, \mathbf{m})$.

\section{Projection-Faithful Reconstruction via Generative Depth Estimation}
Following widely successful diffusion-based approaches~\cite{ho2020}, we model the conditional posterior distribution $p(\mathbf{y} | \mathbf{x}, \mathbf{p}, \mathbf{m})$ using a Latent Diffusion Model (LDM) designed for high-resolution image synthesis~\cite{rombach2022}.

\subsection{Latent Diffusion Framework}
To maintain computational efficiency while preserving fine-grained geometric details, we perform the diffusion process in a compressed latent space. We utilize a variational autoencoder (VAE)~\cite{kingma2019} consisting of an encoder $\mathcal{E}$ and a decoder $\mathcal{D}$. The target depth map $\mathbf{y}$ is encoded into a latent representation $\mathbf{z}_0 = \mathcal{E}(\mathbf{y})$, where $\mathbf{z}_0 \in \mathbb{R}^{h \times w \times c}$ is spatially downsampled by a factor of $f=8$.

The diffusion process is modeled as a Markov chain that incrementally destroys the structure of $\mathbf{z}_0$ by adding Gaussian noise over $T$ timesteps, resulting in a sequence $\mathbf{z}_1, \dots, \mathbf{z}_T$, where $\mathbf{z}_T \sim \mathcal{N}(0, \mathbf{I})$.

The generative process is learned by training a time-conditional denoising network $\epsilon_\theta$ to predict the noise component $\epsilon$ added to a noisy latent $\mathbf{z}_t$, conditioned on the timestep $t$ and our geometric conditions $\mathbf{c} = (\mathbf{x}, \mathbf{p}, \mathbf{m})$. The objective function is the standard noise-prediction loss:
$$\mathcal{L}_{LDM} = \mathbb{E}_{t, \mathbf{z}_0, \epsilon, \mathbf{c}} \left[\| \epsilon - \epsilon_\theta(\mathbf{z}_t, t, \mathbf{c}) \|_2^2 \right]$$

\subsection{Network Architecture and Conditioning}
Our denoising backbone $\epsilon_\theta$ follows the U-Net architecture~\cite{ronneberger2015} used in Stable Diffusion, featuring alternating ResNet blocks and spatial self-attention mechanisms.

\textbf{ControlNet-Style Injection.} Unlike semantic text conditioning, which is typically handled via cross-attention, sketch conditioning requires strict spatial alignment between the input strokes and the output geometry. To enforce this pixel-to-pixel correspondence, we adopt a ControlNet-style conditioning mechanism~\cite{rombach2022, zhang2023}. The ControlNet encoder extracts a set of multi-scale conditional feature maps, collectively denoted as $\{\mathbf{c}_i\}_{i=1}^4$, at various spatial resolutions. These hierarchical representations are then added directly to the corresponding intermediate hidden states of the U-Net decoder (Figure~\ref{fig:architecture}).

\subsection{Conditioning Encoder Variants}
The choice of $\mathcal{E}_{cond}$ is critical for interpreting the sparse, abstract nature of wireframe drawings. Considering the domain shift of line drawings compared to natural images, we investigate three distinct architectures for the conditioning branch:
\begin{enumerate}
    \item \textbf{Latent Encoder (VAE-KL):} We initialize $\mathcal{E}_{cond}$ using the weights of the primary VAE encoder, representing an encoder pre-trained specifically to understand underlying wireframe geometries.
    \item \textbf{Vision Transformer (ViT):} We employ a standard ViT architecture trained from scratch~\cite{dosovitskiy2020}.
    \item \textbf{Pre-trained DinoV2:} We leverage a ViT pre-trained with the DinoV2 objective~\cite{oquab2023}. Despite being trained on natural images, DinoV2 encodes rich, generalized visual representations that surpass what a wireframe-only encoder can initially capture. Thus, we fine-tune this expansive prior specifically to the wireframe domain.
\end{enumerate}

\subsection{Geometric Representation: Fixed-Range Normalized Disparity}
A critical design choice in our reconstruction pipeline is the representation of geometry. Rather than predicting metric depth directly, we predict a normalized disparity field derived from inverse depth. This choice compresses the dynamic range of the target and places all training examples into a consistent bounded interval.

Let $Z_{u,v}$ denote the rendered depth at pixel $(u,v)$. For foreground pixels, we first convert depth to raw disparity via inverse depth, and then normalize it using fixed near and far depth planes, $Z_{\mathrm{near}}$ and $Z_{\mathrm{far}}$. In our implementation, to satisfy global orthographic constraints, these are defined as global constants rather than object-specific bounds. The target is therefore defined as:
$$y_{u,v} = \frac{\frac{1}{Z_{u,v}} - \frac{1}{Z_{\mathrm{far}}}}{\frac{1}{Z_{\mathrm{near}}} - \frac{1}{Z_{\mathrm{far}}}}$$

\section{Dataset Generation}
\label{sec:dataset}
To train a generative model capable of generalizing across diverse engineering wireframes, we construct large-scale datasets of wireframe-depth pairs derived from the ABC Dataset~\cite{koch2019}. We generate two datasets. The first (v1) begins from the filtered ABC subset used by Wang et al.~\cite{wang2022}, which removes unnecessarily complex shapes and duplicates, resulting in 10,076 unique CAD models. We use this subset to maintain alignment with prior wireframe-specific reconstruction work and render each model from multiple orthographic viewpoints to produce over one million paired sketch and depth samples at $256 \times 256$ resolution. This dataset supports our conditioning-encoder and model-scaling study. The second, expanded dataset (v2) is curated directly from the raw ABC corpus with our own filter family (Section~\ref{sec:expanded_curation}) and rendered at $512 \times 512$, producing 2,138,056 pairs across roughly 90,000 unique CAD models, resulting in a $9\times$ increase in shape diversity. This dataset trains our final high-resolution model.

\begin{figure}[ht]
    \centering
    \includegraphics[width=\linewidth]{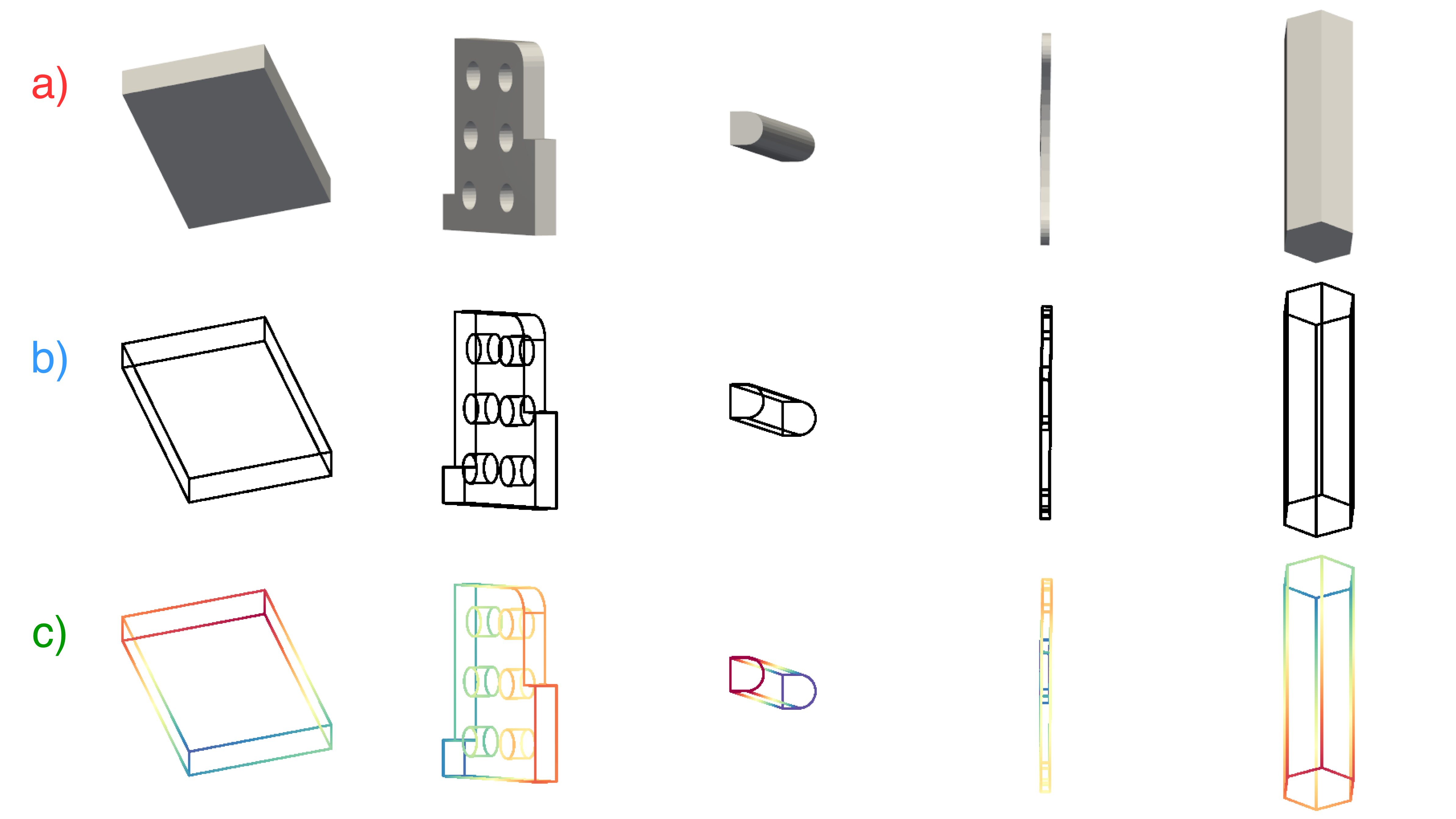} 
    \caption{\textbf{Wireframe generation pipeline.} Our wireframe generation pipeline begins with a) the initial CAD object imported from ABC, into b) the CAD wireframe/sketch mask, and finally c) the resulting depth map of the sketch projection.}
    \label{fig:pipeline}
\end{figure}

\subsection{Rendering Pipeline}
For each valid CAD, we render 2D wireframe projections and corresponding ground truth depth maps by modifying the rasterization pipeline in Wang et al.~\cite{wang2022} built on PythonOCC (Figure~\ref{fig:pipeline}).
\begin{itemize}
    \item \textbf{Camera Model:} We utilize an orthographic projection camera and sample viewpoints uniformly from a hemisphere. In v1, each wireframe was rendered from 100 randomly sampled views. In v2, each wireframe is rendered from 24 random views, relying on shape diversity rather than view diversity. Both datasets are enhanced at training time by dataloader augmentation (Section~\ref{sec:partial_depth}).
    \item \textbf{Stroke Width:} v2 renders strokes at half the physical diameter of v1 while maintaining a raster width of roughly 2 pixels at $512 \times 512$. Thinner physical strokes increase the visual separation between adjacent edges, directly reducing the accidental pixel ratio identified as a primary error driver in our analysis.
    \item \textbf{Depth Maps:} Ground truth depth maps are generated via Z-buffering.
\end{itemize}
Example renders at each stage of the pipeline are shown in Figure~\ref{fig:pipeline}.

\subsection{Expanded Shape Curation}
\label{sec:expanded_curation}
For the expanded dataset, we replace the fixed Wang et al. subset with a streaming curation pipeline over the raw ABC chunks: each chunk is downloaded, inventoried, filtered on shape metadata, deduplicated against previously accepted shapes, and rendered before moving to the next chunk. The filters target moderate complexity with geometric variety, allowing a single root solid, at most 96 edges and 192 faces, and caps on the short-edge fraction and normalized total edge length. We avoid penalizing curved geometry, as heavily curved and organic shapes are where depth-based reconstruction is most valuable.

This curation philosophy diverges from how prior engineering-sketch systems construct training data. CAD methods synthesize training corpora from sketch-and-extrude sequences whose profile vocabulary is limited to lines, arcs, and circles~\cite{wu2021, li2022}. Free-form B-spline geometry is unrepresentable in the data by construction. Topology-based methods filter ABC to shapes that satisfy their structural assumptions. The subset of Wang et al.~\cite{wang2022}, which our v1 dataset adopts, retains manifold shapes with valid face loops and modest complexity. Table~\ref{tab:dataset_comparison} identifies the statistical differences: our expanded dataset doubles the structural complexity (median 36 vs.\ 18 edges per shape; maximum 96 vs.\ 53) while retaining curve-rich and free-form geometry that prior filtering excludes. The complexity and curves our benchmark includes are exactly those that prior training distributions exclude or under-represent.

\begin{table}[t]
\centering
\small
\renewcommand{\arraystretch}{1.2}
\caption{\textbf{Training-corpus comparison: prior-work ABC subset (v1) vs.\ our expanded curation (v2).} All statistics are computed with the same measurement code on both datasets. The expanded dataset doubles admissible structural complexity, extending into regimes excluded by prior filtering.}
\label{tab:dataset_comparison}
\resizebox{\linewidth}{!}{%
\begin{tabular}{l c c}
\toprule
& \textbf{v1 (Wang et al.\ subset)} & \textbf{v2 (ours)} \\
\midrule
CAD shapes & 10,076 & 90,951 \\
Rendered pairs & 1.00M & 2.14M \\
Resolution & $256 \times 256$ & $512 \times 512$ \\
Edges per shape (median / p90 / max) & 18 / 39 / 53 & 36 / 77 / 96 \\
\bottomrule
\end{tabular}%
}
\end{table}

\subsection{Simulating Partial Depth}
\label{sec:partial_depth}
A key contribution of our method is the ability to reconstruct from partial depth. To train this capability, we must simulate realistic intermediate states of a drawing. An intermediate state has two components: which strokes the user has drawn so far, and which of those strokes already carry depth from a previous reconstruction step. We simulate both on the shared edge-vertex graph $\mathcal{G} = (V,E)$ of the visible wireframe.

For the depth component, naive random masking (e.g., dropping 50\% of pixels) fails to capture human drawing behavior, as users typically draw connected strokes rather than scattered pixels. We instead propose a graph-based masking strategy:
\begin{enumerate}
    \item For 50\% of training samples, we select a random ``start node'' $v_{start} \in V$.
    \item We perform a breadth-first search (BFS) traversal starting from $v_{start}$.
    \item We retain the depth values for all edges visited by the BFS until a random threshold $k \sim U[10\%, 90\%]$ of the total wireframe is covered.
\end{enumerate}

For the stroke component, we apply a similar shape-level augmentation on the same graph. A random subset of edges is removed entirely, simulating a drawing whose geometry is itself incomplete rather than merely lacking depth. In the expanded dataset, both mechanisms are applied in the dataloader so every epoch sees fresh intermediate states of every drawing.

\subsection{Dataset Statistics}
The v1 dataset consists of 1,004,051 unique pairs from 10,076 shapes, a magnitude increase over the approximately 10,076 samples used in previous wireframe-specific works~\cite{wang2022}. The expanded v2 dataset consists of 2,138,056 unique pairs at $512 \times 512$ resolution from roughly 90,000 curated shapes, increasing shape diversity by another order of magnitude.

\section{Experimental Setup}
\subsection{Implementation Details}
We implement our framework using PyTorch~\cite{paszke2019} and the Hugging Face library~\cite{wolf2019}, leveraging the latter to integrate pre-trained backbones (e.g., DinoV2) for the conditioning encoder. Models in the conditioning-encoder and scaling study are trained on the v1 dataset at a spatial resolution of $256 \times 256$ pixels. Our final model is trained on the expanded v2 dataset at $512 \times 512$.

We utilize the AdamW optimizer~\cite{loshchilov2017} for training. To preserve the learned features of pre-trained encoders (such as DinoV2 or the VAE-KL encoder), we apply a differential learning rate strategy: the learning rate for the pre-trained conditioning encoder is set one order of magnitude lower than that of the randomly initialized denoising U-Net.

Training of the v1 models was performed on a cluster of $4 \times$ NVIDIA RTX 4090 GPUs, with a batch size of 192 for 100 epochs. Total training time was about 128 GPU hours for smaller models (VAE and small ViTs) and about 336 hours for the larger scale U-Nets with the DinoV2-Base configuration. For our largest v1 model, we upgraded to a $4 \times$ NVIDIA RTX 5090 rig, with a total training time of 240 GPU hours.

Migrating to $512 \times 512$ required retraining both stages of the latent pipeline: the retrained VAE removes the reconstruction ceiling of the original autoencoder (Section~\ref{sec:resolution_scaling}), and because the latent space changes, the U-Net is retrained from scratch rather than fine-tuned. The model is trained on $4 \times$ NVIDIA H100 GPUs at an effective batch size of 200 for 25 epochs (approximately 310 GPU hours).

\subsection{Benchmarking Setup \& Evaluation Metrics}

Because our model outputs normalized depth over visible sketch pixels, we evaluate reconstruction quality with depth metrics. This output contract differs from CAD-command reconstruction, face-loop recovery, mesh completion, and sketch-to-asset generation, whose metrics typically measure command accuracy, topology recovery, Chamfer distance, IoU, or perceptual quality~\cite{li2020, li2022, wang2022, tono2026}. Direct metric-to-metric comparison is therefore meaningful only when output spaces are compatible or can be converted without adding additional assumptions. Converting a CAD program to depth (or depth to a CAD program) scores one contract with the other's metric and favors whatever contract is being compared on. Thus, only against monocular depth estimators are we able to compare numerically (Table~\ref{tab:metrics}, Figure~\ref{fig:off_the_shelf}). Against engineering reconstruction systems with different contracts, we can only compare on assumption coverage: Table~\ref{tab:capability} records, from each method's own stated assumptions, which inputs it admits and which outputs it produces.

\begin{table*}[t!]
\centering
\sffamily
\small
\renewcommand{\arraystretch}{1.25}
\caption{\textbf{Assumption comparison across output contracts.} Marks are derived from each method's stated input requirements and output representation (\checkmark~supported, $\sim$~partially supported, $\times$~unsupported by the method's assumptions). Our benchmark corpus contains geometrically freeform edges expressible by neither a line nor a circular arc (``curves beyond fixed vocabulary'') and is curated without manifold or closed-face requirements (``open/partial wireframes''). Depth Anything V2 shares our contract and is compared numerically in Table~\ref{tab:metrics}.}
\label{tab:capability}
\resizebox{\textwidth}{!}{%
\begin{tabular}{l c c c c c c c}
\toprule
\textbf{Method}
& \textbf{\shortstack{Raster input\\(no stroke order /\\vectorization)}}
& \textbf{\shortstack{Curves beyond\\fixed vocabulary}}
& \textbf{\shortstack{Open / partial\\wireframes}}
& \textbf{\shortstack{Preserves drawn\\projection}}
& \textbf{\shortstack{Iterative\\partial-3D\\conditioning}}
& \textbf{\shortstack{Metric 3D\\output}}
& \textbf{\shortstack{Editable CAD\\output}} \\
\midrule
Lipson \& Shpitalni~\cite{lipson1996} & $\times$ & $\times$ & $\times$ & \checkmark & $\times$ & \checkmark & $\times$ \\
Sketch2CAD~\cite{li2020} & $\times$ & $\times$ & $\times$ & $\times$ & \checkmark & \checkmark & \checkmark \\
DeepCAD~\cite{wu2021} & $\times$ & $\times$ & $\times$ & $\times$ & $\times$ & \checkmark & \checkmark \\
Free2CAD~\cite{li2022} & $\times$ & $\sim$ & $\times$ & $\times$ & \checkmark & \checkmark & \checkmark \\
Neural Face ID~\cite{wang2022} & $\times$ & $\sim$ & $\times$ & \checkmark & $\times$ & $\sim$ & $\times$ \\
Depth Anything V2~\cite{yang2024v2} & \checkmark & \checkmark & \checkmark & \checkmark & $\times$ & \checkmark & $\times$ \\
\textbf{Ours} & \checkmark & \checkmark & \checkmark & \checkmark & \checkmark & \checkmark & $\times$ \\
\bottomrule
\end{tabular}%
}
\end{table*}

We evaluate on two held-out test suites, one per dataset. For the v1 study, we use the 504 shapes inside our dedicated test dataset and render four isometric viewpoints per shape, resulting in a benchmarking dataset of 2016 unique viewpoints. To quantitatively evaluate the geometric accuracy of the reconstructed depth maps, we report standard depth estimation metrics. To account for the stochastic nature of the diffusion process, we generate five predictions per sample, reporting both the average and the best-case performance. Let $y_i$ denote the ground truth normalized disparity and $\hat{y}_i$ denote the predicted disparity at pixel $i$, for a total of $N$ valid pixels.

1. \textbf{Mean Absolute Error (MAE):} This metric measures the raw magnitude of errors in a set of predictions in units of globally normalized disparity.
$$\text{MAE} = \frac{1}{N} \sum_{i=1}^{N} |y_i - \hat{y}_i|$$

2. \textbf{Normalized MAE (NMAE):} This metric normalizes the MAE to each image's disparity range. Most shapes would not fill the global disparity bound; thus, this metric is necessary for per image understanding.
$$\text{MAE}_{\text{normalized}} = \frac{\text{MAE}}{\max(y) - \min(y) + \epsilon}$$

3. \textbf{Absolute Relative Error (AbsRel):} This metric normalizes the error relative to the ground truth value, penalizing errors more heavily in regions of lower disparity (further depth).
$$\text{AbsRel} = \frac{1}{N} \sum_{i=1}^{N} \frac{|y_i - \hat{y}_i|}{y_i}$$

4. \textbf{Threshold Accuracy ($\delta$):} We measure the percentage of pixels where the ratio between the predicted and ground truth values is within a specific threshold. We report the standard metric $\delta < 1.25$:
$$\delta = \frac{1}{N} \sum_{i=1}^{N} \mathbb{I} \left(\max \left(\frac{\hat{y}_i} {y_i}, \frac{y_i} {\hat{y}_i} \right) < 1.25 \right)$$
where $\mathbb{I}(\cdot)$ is the indicator function.

\textbf{Depth-reversal-aware scoring.} Under orthographic projection, every line drawing admits two equally valid depth interpretations related by reflecting all depths about the scene center. Thus, we therefore score each prediction against both the annotated ground truth and its depth-reversed counterpart, and report the metrics of the better-matching interpretation.

\begin{figure}[ht]
    \centering
    \includegraphics[width=\linewidth]{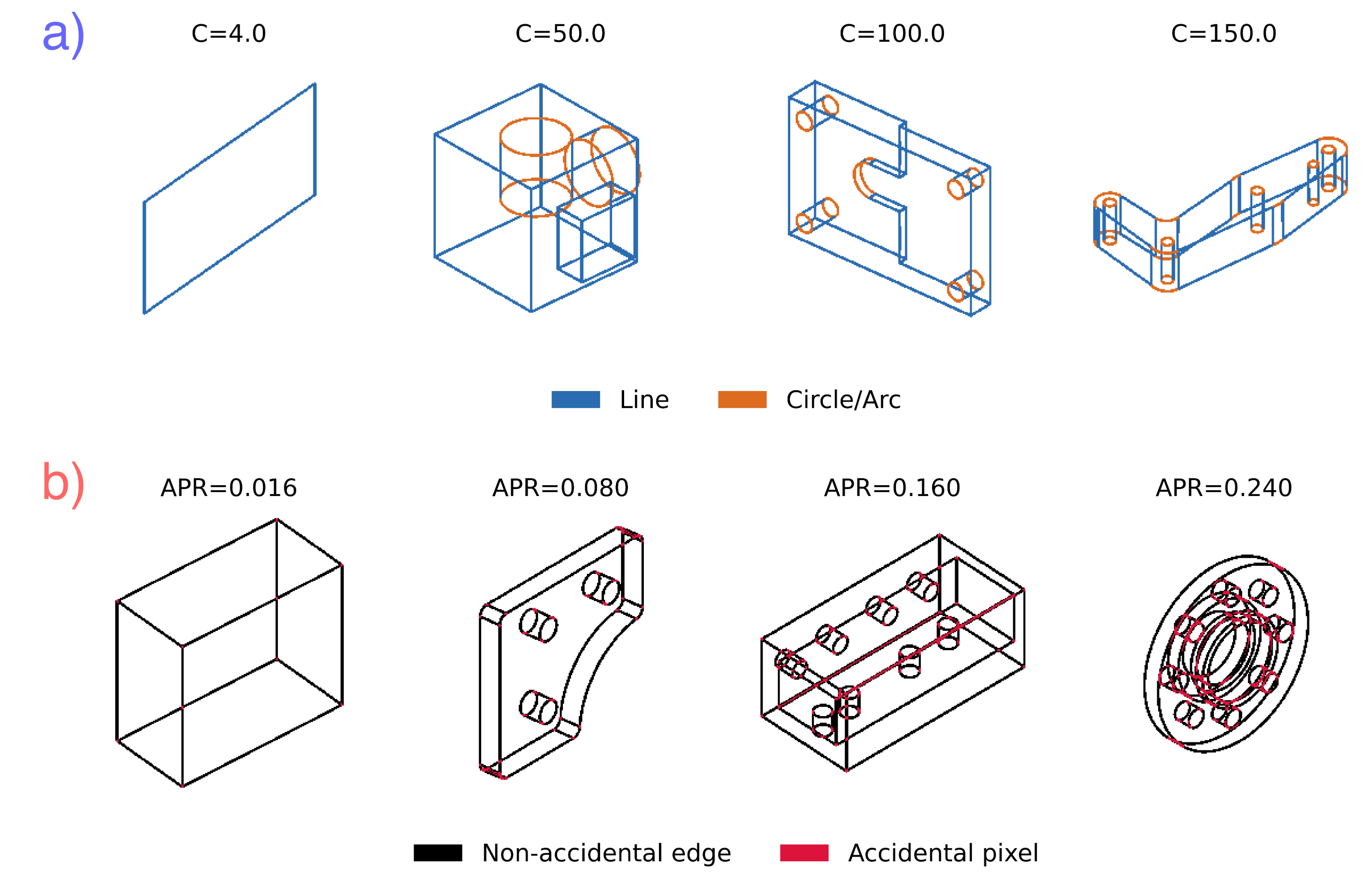} 
    \caption{\textbf{Example shapes of various a) shape complexities and b) APRs.} To understand the effect of various measurements of shape complexity on model performance, we segment our dataset based on these parameters. For shape complexity, we color code by the primitive type, and for APR, we mark accidental pixel regions as red.}
    \label{fig:apr_complexity_ex}
\end{figure}

In total, we run experiments covering each conditioning encoder (VAE-KL, ViT small scratch, DinoV2). Then, we perform additional ablations studies on the DinoV2 model by halving the dataset size and testing the newer DinoV3 encoder ~\cite{simeoni2025}. Lastly, to determine how the model reacts to compute scaling, we provide DinoV2 and V3 base models with 3$\times$ and 6$\times$ larger variants.  

In addition, to gauge how the model performs against shapes of various complexity, we provide two additional parameters: accidental pixel ratio (APR) and curve complexity. APR is computed as the proportion of all foreground pixels that occlude one or more components, while curve complexity ratio is a summation of the per-edge primitive costs. To keep calculations simple and robust to potentially unclean CAD data, we assign a score of 1 to lines and a score of 3 to non-lines. Figure~\ref{fig:apr_complexity_ex} displays example shapes of various shape complexities and APRs.

\section{Results \& Discussion}
\subsection{Comparison with Baseline and Off-the-Shelf Approaches}
To evaluate the efficacy of our generative diffusion-based approach, we first compare our pipeline against deterministic models and state-of-the-art, off-the-shelf depth and 3D reconstruction frameworks.

\textbf{Deterministic Regression:} As established, the orthographic projection of wireframes creates an inherently ill-posed problem, as a single 2D sketch can mathematically correspond to multiple valid 3D interpretations. When we attempted to train a standard deterministic regression model to map $x \rightarrow y$, the network consistently failed to resolve these ambiguities, instead converging to the mean of all possible modes.

\begin{table*}[htbp]
\centering
\renewcommand{\arraystretch}{1.3}
\caption{\textbf{Quantitative benchmark results.} The upper panel reports the conditioning-encoder and scaling study: models trained on the v1 dataset at $256 \times 256$ and evaluated on the v1 test suite. The lower panel reports the expanded benchmark: models evaluated on the v2 dataset. Numbers are comparable within a panel, not across panels. The VAE floor rows report each autoencoder's round-trip reconstruction error on ground-truth depth maps, forming a hard lower bound on any diffusion model operating in that latent space; the $256$ system's best-case error sits at its floor, while the new 512 VAE removes this ceiling and the fully expanded model converts it into a roughly $2\times$ reduction in best-case error.}
\label{tab:metrics}
\resizebox{\textwidth}{!}{%
\begin{tabular}{l cccc | ccc | ccc}
\toprule
& \multicolumn{4}{c}{\textbf{Model Details}} & \multicolumn{3}{c}{\textbf{Average Prediction}} & \multicolumn{3}{c}{\textbf{Best Prediction}} \\
\cmidrule(lr){2-5} \cmidrule(lr){6-8} \cmidrule(lr){9-11}
\textbf{Model} & \textbf{Params} & \textbf{Data} & \textbf{Res.} & \textbf{GPU Hrs} & \textbf{NMAE} & \textbf{AbsRel} & \textbf{$\delta < 1.25$} & \textbf{NMAE} & \textbf{AbsRel} & \textbf{$\delta < 1.25$} \\
\midrule
\multicolumn{11}{l}{\textit{v1 benchmark (10k shapes dataset)}} \\
\midrule
Baseline (average) & - & - & - & - & $0.221 \pm 0.001$ & $0.262 \pm 0.002$ & $0.499 \pm 0.004$ & $0.221 \pm 0.001$ & $0.262 \pm 0.002$ & $0.499 \pm 0.004$ \\
Depth Anything V2~\cite{yang2024v2} & 98M & - & 256 & - & $0.172 \pm 0.002$ & $0.208 \pm 0.002$ & $0.646 \pm 0.006$ & $0.172 \pm 0.002$ & $0.208 \pm 0.002$ & $0.646 \pm 0.006$ \\
VAE encoder & 63M & 1M & 256 & 112 & $0.101 \pm 0.002$ & $0.115 \pm 0.002$ & $0.863 \pm 0.005$ & $0.057 \pm 0.002$ & $0.065 \pm 0.002$ & $0.962 \pm 0.003$ \\
ViT small scratch & 84M & 1M & 256 & 128 & $0.100 \pm 0.002$ & $0.114 \pm 0.002$ & $0.868 \pm 0.005$ & $0.057 \pm 0.002$ & $0.065 \pm 0.002$ & $0.960 \pm 0.003$ \\
DinoV2 small & 84M & 500K & 256 & 128 & $0.072 \pm 0.002$ & $0.081 \pm 0.002$ & $0.927 \pm 0.004$ & $0.043 \pm 0.001$ & $0.047 \pm 0.002$ & $0.978 \pm 0.002$ \\
DinoV2 small & 84M & 1M & 256 & 128 & $0.069 \pm 0.002$ & $0.077 \pm 0.002$ & $0.941 \pm 0.003$ & $0.041 \pm 0.001$ & $0.047 \pm 0.001$ & $0.983 \pm 0.002$ \\
DinoV3 small & 84M & 1M & 256 & 128 & $0.070 \pm 0.002$ & $0.079 \pm 0.002$ & $0.937 \pm 0.004$ & $0.042 \pm 0.001$ & $0.048 \pm 0.002$ & $0.981 \pm 0.002$ \\
DinoV2 base & 227M & 1M & 256 & 336 & $0.059 \pm 0.002$ & $0.065 \pm 0.002$ & $0.956 \pm 0.003$ & $0.036 \pm 0.001$ & $0.039 \pm 0.001$ & $0.987 \pm 0.001$ \\
DinoV3 base & 227M & 1M & 256 & 336 & $0.066 \pm 0.002$ & $0.074 \pm 0.002$ & $0.940 \pm 0.004$ & $0.038 \pm 0.001$ & $0.043 \pm 0.001$ & $0.984 \pm 0.002$ \\
\textbf{DinoV2 base XL} & \textbf{496M} & \textbf{1M} & \textbf{256} & \textbf{240} & $\mathbf{0.053} \pm \mathbf{0.002}$ & $\mathbf{0.058} \pm \mathbf{0.002}$ & $\mathbf{0.960} \pm \mathbf{0.003}$ & $\mathbf{0.031} \pm \mathbf{0.001}$ & $\mathbf{0.034} \pm \mathbf{0.001}$ & $\mathbf{0.989} \pm \mathbf{0.001}$ \\
DinoV3 base XL & 496M & 1M & 256 & 240 & $0.067 \pm 0.002$ & $0.075 \pm 0.002$ & $0.937 \pm 0.004$ & $0.042 \pm 0.001$ & $0.046 \pm 0.002$ & $0.980 \pm 0.002$ \\
\midrule
\multicolumn{11}{l}{\textit{Expanded v2 benchmark (90k shape dataset)}} \\
\midrule
DinoV2 base XL (no fine-tune) & 496M & v1 (1M) & 256 & - & $0.121 \pm 0.001$ & $0.123 \pm 0.001$ & $0.850 \pm 0.002$ & $0.091 \pm 0.001$ & $0.093 \pm 0.001$ & $0.920 \pm 0.002$ \\
DinoV2 base XL (fine tuned) & 496M & v2 (2.1M) & 256 & 240 & $0.097 \pm 0.002$ & $0.095 \pm 0.001$ & $0.925 \pm 0.001$ & $0.072 \pm 0.001$ & $0.071 \pm 0.001$ & $0.965 \pm 0.001$ \\
\textbf{DinoV2 base XL high-res} & \textbf{638M} & \textbf{v2 (2.1M)} & \textbf{512} & \textbf{310} & $\mathbf{0.070} \pm \mathbf{0.001}$ & $\mathbf{0.076} \pm \mathbf{0.001}$ & $\mathbf{0.932} \pm \mathbf{0.001}$ & $\mathbf{0.039} \pm \mathbf{0.001}$ & $\mathbf{0.040} \pm \mathbf{0.001}$ & $\mathbf{0.982} \pm \mathbf{0.001}$ \\
\textit{VAE floor (256 latent space)} & - & - & 256 & - & $0.069$ & $0.069$ & $0.979$ & $0.069$ & $0.069$ & $0.979$ \\
\textit{VAE floor (512 latent space)} & - & - & 512 & - & $0.002$ & $0.003$ & $1.000$ & $0.002$ & $0.003$ & $1.000$ \\
\bottomrule
\end{tabular}%
}
\end{table*}

\textbf{Comparison with Off-the-Shelf Frameworks:} We further trialed our wireframe inputs against leading zero-shot vision models, which share enough of our output contract for numeric and visual comparison.

\begin{figure}[ht]
    \centering
    \includegraphics[width=\linewidth]{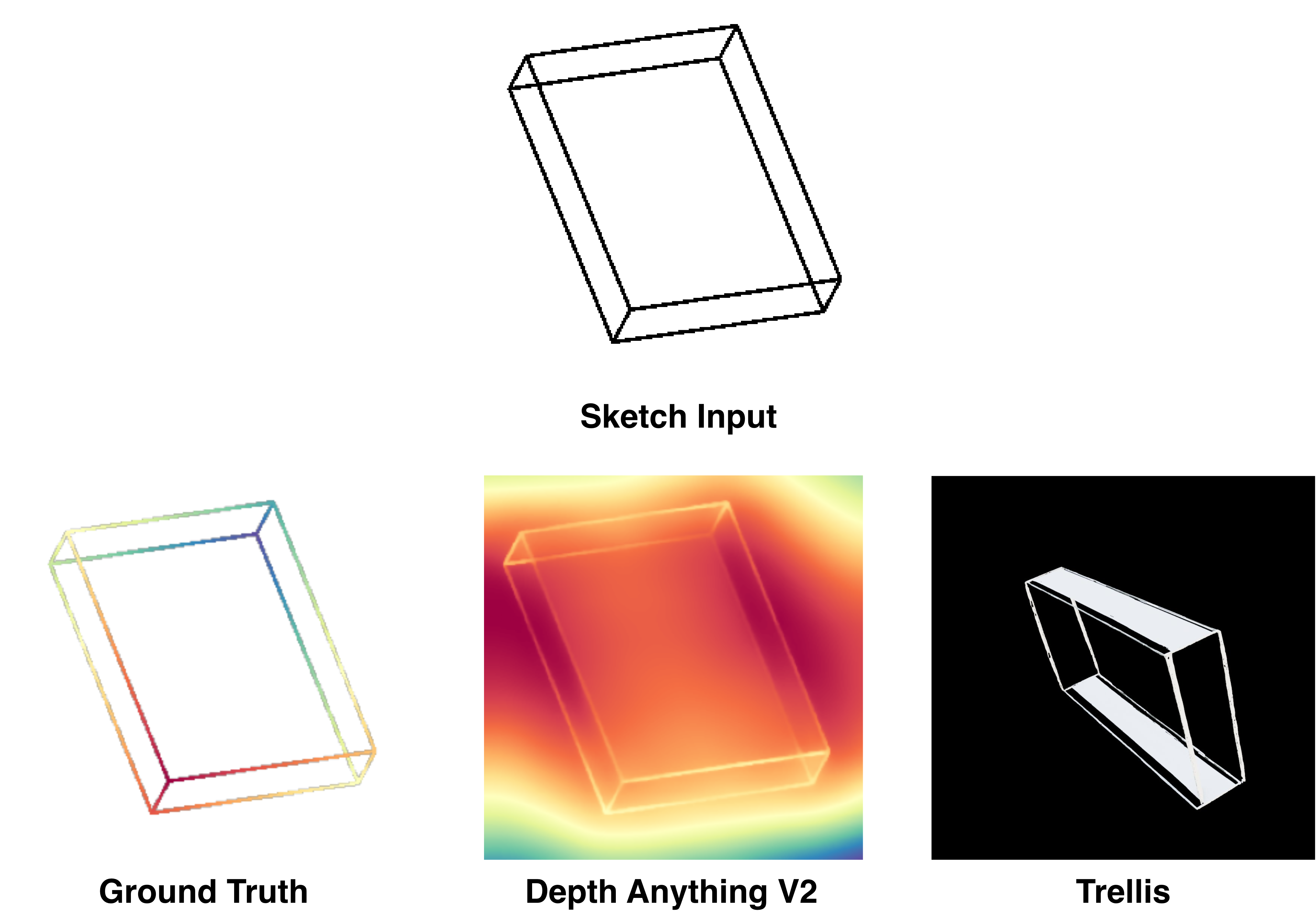} 
    \caption{\textbf{Failures of off-the-shelf models.} For a simple rectangular prism, off-the-shelf models fail in both the depth estimation task (as predicted by Depth Anything V2) and direct 3D reconstruction (Trellis).}
    \label{fig:off_the_shelf}
\end{figure}

Initially, we attempted to use an off-the-shelf depth estimator, namely the widely successful Depth Anything V2~\cite{yang2024v2}. While Depth Anything understands basic semantics of near and far regions, it is clearly unsuited to the unique visual domain of wireframe projections (Figure~\ref{fig:off_the_shelf}).

Additionally, we performed inference on Microsoft's Trellis~\cite{xiang2025} in an attempt to generate a 3D wireframe asset directly. Not only was the generation process too slow to support our intended ``sketch-reconstruct-sketch'' pipeline, but the results exhibited severe geometric distortions, underscoring a clear input unsuitability (Figure~\ref{fig:off_the_shelf}).

\textbf{Comparison with Engineering Reconstruction Systems:} For engineering-sketch systems with different output contracts, numeric comparison is undefined (Table~\ref{tab:capability}); the comparison that is well-defined is what geometry each system demonstrates and what geometry its training distribution admits. The ``Shape'' column of Table~\ref{tab:related_works} places representative geometry from each closely related system's demonstrated domain alongside a reconstruction of ours. The prior domains concentrate on prismatic, low-curve-count geometry, consistent with their line/arc/circle program vocabularies and complexity-filtered training subsets (Table~\ref{tab:dataset_comparison}), whereas our method faithfully reconstructs a 60-stroke sketch in which 34 strokes are free-form B-spline curves from a single raster input. This is the practical consequence of the contract differences in Table~\ref{tab:capability}: the geometry regime where engineering sketches are hardest is the regime our formulation is designed to cover.

\subsection{Hyperparameter Analysis: Conditioning Encoders, Dataset Size, and Model Size}
The choice of the conditioning encoder ($\mathcal{E}_{cond}$) dictates how effectively the U-Net denoiser can interpret the sparse, abstract nature of wireframe drawings. We tested three distinct encoder architectures: a Latent Encoder (VAE-KL), a ViT small trained from scratch, and a pre-trained DinoV2 backbone (Table~\ref{tab:metrics}).

Quantitative results (Table \ref{tab:metrics}) indicate that the DinoV2 encoder achieved the best performance across all metrics. While the simpler, CNN-based VAE-KL encoder displayed an initially faster convergence speed than the DinoV2 encoder, DinoV2 eventually reached a lower final loss. This suggests that although modern foundation models are trained primarily on natural images, their learned representations translate to wireframe domains given sufficient fine-tuning. However, DinoV2's superior performance does not necessarily imply that a natural image prior is strictly better than a purely geometric prior, as architectural differences (ViT vs. CNN) play a significant role. Rather, the VAE-KL encoder's rapid initial convergence points to the ideal necessity of a wireframe-specific ViT backbone—one conditioned first on natural images and subsequently fine-tuned on geometric wireframes. Indeed, the fine-tuned DinoV2 encoder utilized in this work fills precisely this gap, representing a preliminary step toward a wireframe-specific vision foundation model.

Ultimately, while encoders trained on natural images provide a suitable foundation, they are limited in the long run. Even as we ran a benchmark with the newer DinoV3, trained on nearly 12$\times$ the amount of data compared to the V2, it resulted in virtually no improvement in the small and base models and a significant deterioration in the base XL. Thus, to improve performance in wireframe specific tasks, it is not enough to simply build a better natural image-based model; it is instead required to build better wireframe specific foundation models. 

\begin{figure}[t]
    \centering
    \includegraphics[width=\linewidth]{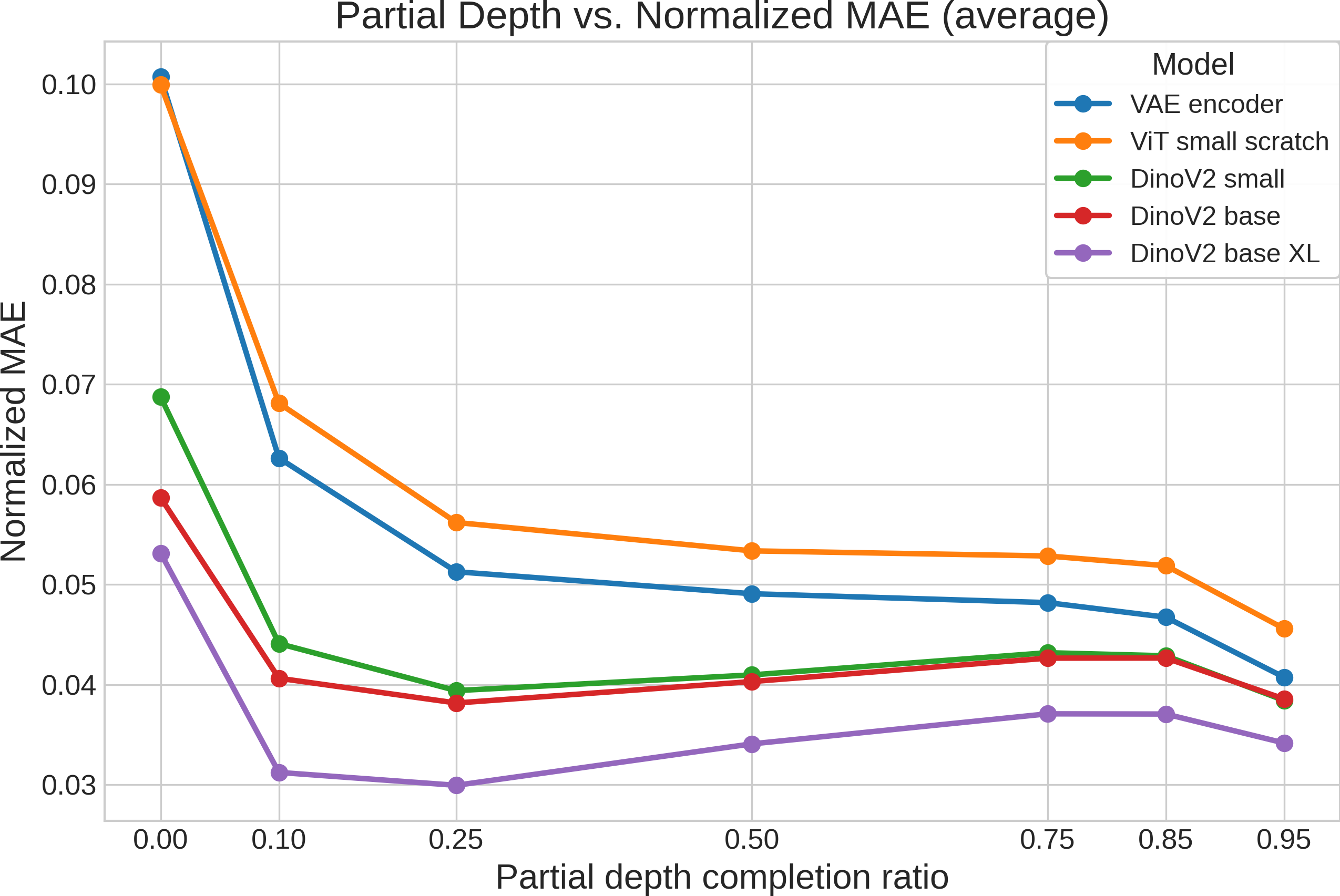}
    \caption{\textbf{Partial depth vs. Normalized MAE (average aggregation) for various encoders.} Overall, even minimal partial depth (10\%--25\%) provides a massive boost to reconstruction accuracy. However, increasing partial depth beyond this yields diminishing returns until reaching the final 95\% threshold.}
    \label{fig:partial_depth}
\end{figure}

Furthermore and predictably, increasing model size improved performance while decreasing dataset size harmed performance. This trend is consistent with the scaling behavior commonly observed in large generative models~\cite{bi2024}. Whereas approaches dependent on stroke primitives must be manually adapted and retrained for new curve topologies, our approach offers a straightforward path for continuous improvement via increased computational capacity, either in parameter size or dataset size.

\begin{figure*}[t]
    \centering
    \includegraphics[width=\textwidth]{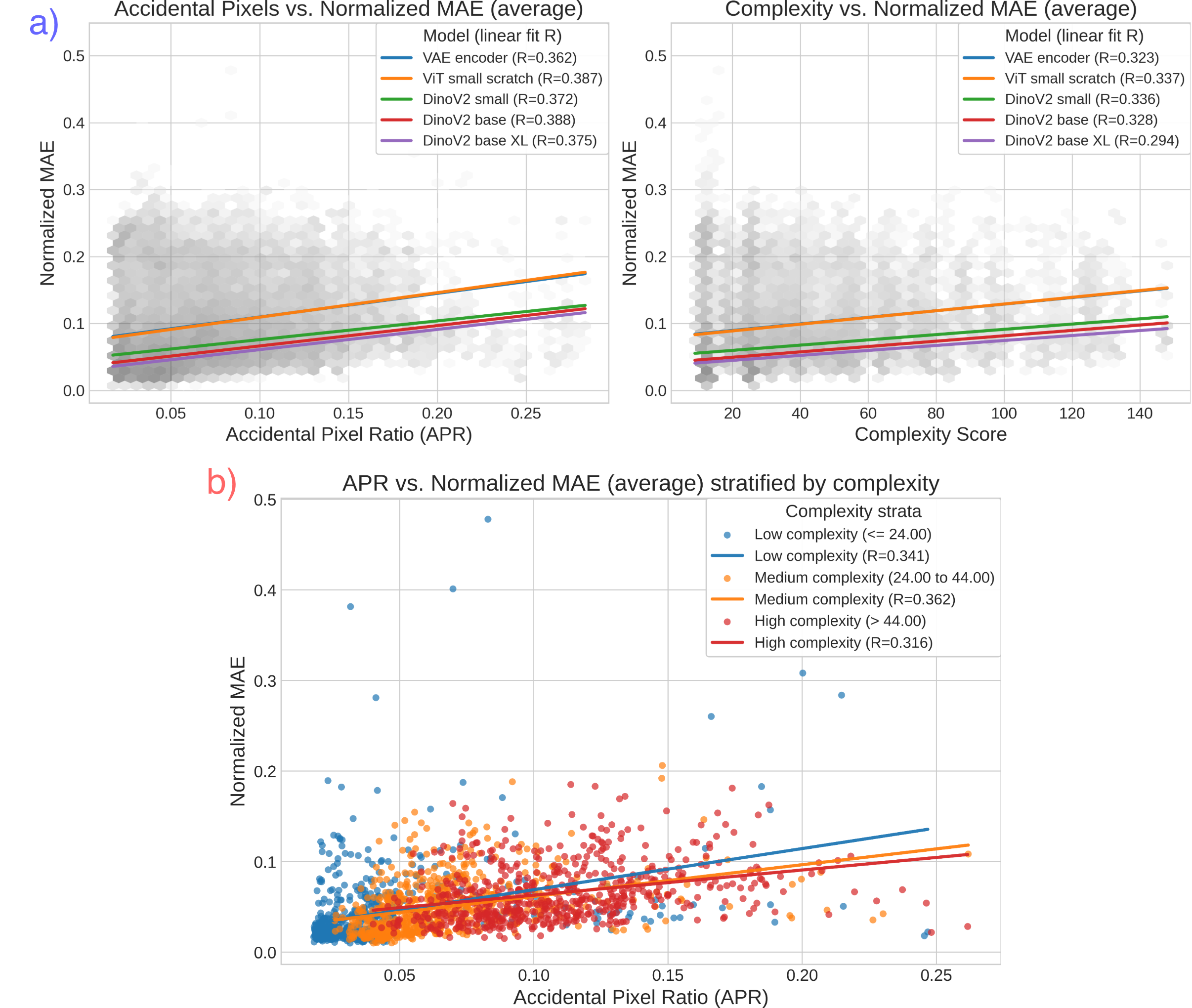} 
    \caption{\textbf{APR and complexity analysis.} a) APR \& shape complexity score vs. normalized MAE for various encoders. Each plot displays density bins pooled across predictions of all models with x-axis outliers removed and b) APR vs. normalized MAE binned by shape complexity for the best DinoV2 model. While APR and complexity are both correlated with increasing error, reducing APR provides a simple path to improvement even for complex shapes.}
    \label{fig:complexity}
\end{figure*}

\subsection{Evaluation of Partial Depth and Sequential Sketching}
A key contribution of our method is the ability to intuitively reconstruct from partial depth, enabling a ``sketch-reconstruct-sketch'' workflow. To validate this, we evaluated the v1 models' performance while varying the partial depth threshold $k \sim U[10\%, 90\%]$ using our graph-based BFS masking strategy. 

As the percentage of provided partial depth ($k$) increases, the NMAE predictably decreases, with every model approaching zero error as the provided partial depth nears completion (Figure~\ref{fig:partial_depth}). However, this decrease is non-linear. Providing the depth of just 10\% of the sketch acts as a geometric ``anchor'' that sufficiently boosts performance, while providing 25\% of the sketch achieves similar performance to providing 75\%.

However, between the 25\% and 85\% thresholds, there exists an interesting plateau and even slight degradation in performance. This plateau suggests that once the global structural details are anchored (accounted for in the first 10\% to 25\%), remaining errors are likely concentrated on smaller structural details that are objectively more difficult to predict, such as a disjointed chunk of the wireframe's edges.

These results show that partial depth cues provide substantial improvement during iterative reconstruction, validating the success of the ``sketch-reconstruct-sketch'' workflow. Supporting iterative sketching not only allows the user to comfortably tackle complex objects in stages, but the iterative geometric anchoring ensures the model autoregressively avoids compounding structural mistakes.

The partial depth curves also illuminate an interesting finding. The first jump in model size (from 84M to 227M) was marked primarily by an increase in zero-shot performance (0\% completion ratio), while the second jump in model size (from 227M to 496M) was marked primarily by increases in non-zero-shot tasks where there was existing partial depth.

\subsection{Robustness to Complexity}
To test the robustness of our pipeline against sketches of varying difficulty, we segmented the v1 test set by shape complexity and APR (Figure~\ref{fig:complexity}).

There exists a positive relationship between predictive error and both shape complexity ($R \approx 0.323$--$0.337$) and APR ($R \approx 0.362$--$0.388$) (Figure~\ref{fig:complexity}a), implying the model is understandably prone to both parameters. However, while shape complexity is constant and unchangeable, APR can be decreased and mediated by strategies such as drawing from better viewpoints, increasing input resolution, and lowering stroke thickness.

The APR and the complexity of the shape are naturally correlated; thus, the question becomes whether reducing the APR can lower the error for shapes of various complexities. If so, the strategies presented provide a straightforward way to improve the model. Thus, in Figure~\ref{fig:complexity}b, we show that regardless of the low / medium / high complexity shapes, the APR and the error continue to be correlated; therefore, by reducing the APR even for the high complexity shapes, we can effectively reduce the error.

Ultimately, while the complexity of a shape is a main driver for error, APR makes up a significant portion of why complexity is such a significant cause for high error.

\subsection{Scaling Resolution and Data}
\label{sec:resolution_scaling}

The preceding analyses identified accidental pixel density as one structural bottleneck of the $256 \times 256$ system. Measuring the autoencoder's round-trip reconstruction error on ground-truth depth maps reveals a second: the $256$ system's best-case error sits essentially on its VAE reconstruction floor, so no amount of further diffusion training could improve it. This motivates the full 512-resolution system of Section~\ref{sec:dataset}: the expanded curve-rich dataset, a retrained VAE with an order-of-magnitude lower floor, and a U-Net retrained from scratch in the new latent space.

The lower panel of Table~\ref{tab:metrics} compares three systems on the expanded v2 benchmark under identical protocol, comparing the original DinoV2 base XL trained on v1, the same architecture fine-tuned on the expanded v2 data at $256 \times 256$, and our 512-resolution model.

Two observations follow. First, expanding the training data alone helps, but the $256$ system's best-case accuracy converges to its VAE reconstruction floor. Second, replacing the latent space and doubling resolution converts that headroom into an approximately $2\times$ reduction in best-of-five error across all metrics, with the largest relative gains exactly where the APR analysis predicted: dense, curve-heavy structures that were previously unresolvable at $256 \times 256$.

\begin{figure*}[t!]
    \centering
    \setlength{\tabcolsep}{6pt}

    \begin{tabular}{c | c | c | c | c | c}
        \toprule
        \textbf{\shortstack{Input\\2D Sketch}} &
        \textbf{\shortstack{Predicted\\Depth}} &
        \textbf{\shortstack{3D Reconstruction \\ \scriptsize (Top: Point Cloud, \\ \scriptsize Bottom: Wireframe)}} &
        \textbf{\shortstack{Input\\2D Sketch}} &
        \textbf{\shortstack{Predicted\\Depth}} &
        \textbf{\shortstack{3D Reconstruction \\ \scriptsize (Top: Point Cloud, \\ \scriptsize Bottom: Wireframe)}} \\
        \midrule

        \raisebox{-.5\height}{\includegraphics[width=0.07\textwidth]{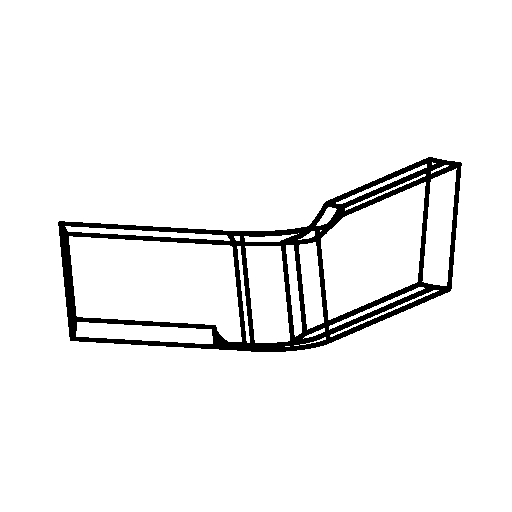}} &
        \raisebox{-.5\height}{\includegraphics[width=0.07\textwidth]{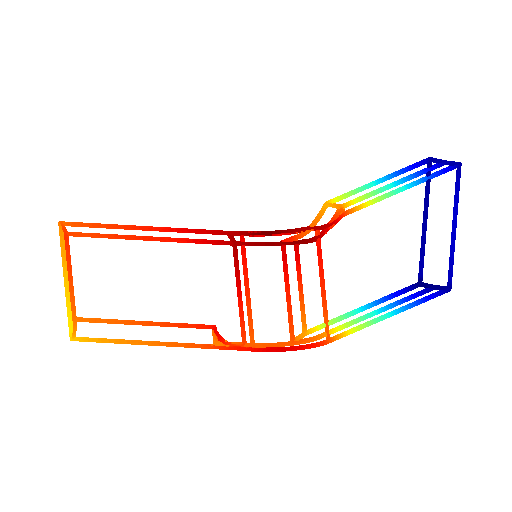}} &
        \raisebox{-.5\height}{\includegraphics[width=0.22\textwidth]{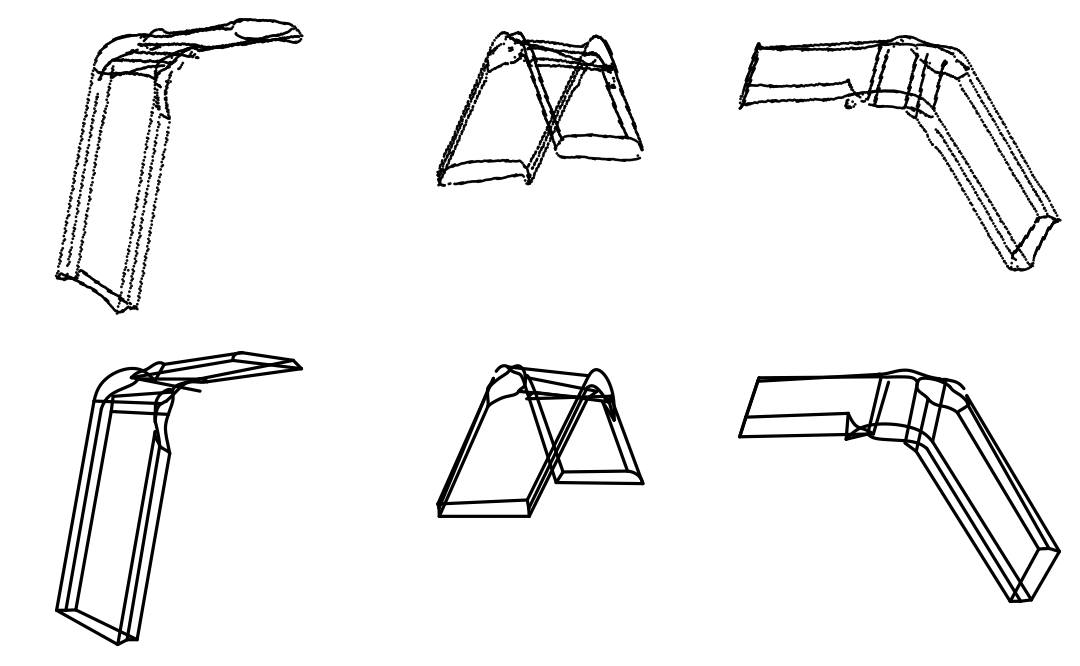}} &
        \raisebox{-.5\height}{\includegraphics[width=0.07\textwidth]{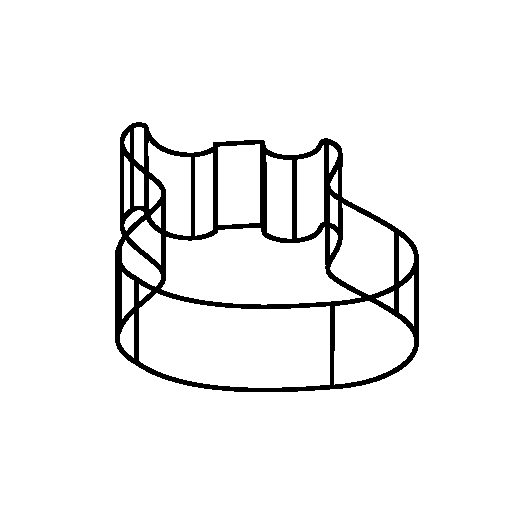}} &
        \raisebox{-.5\height}{\includegraphics[width=0.07\textwidth]{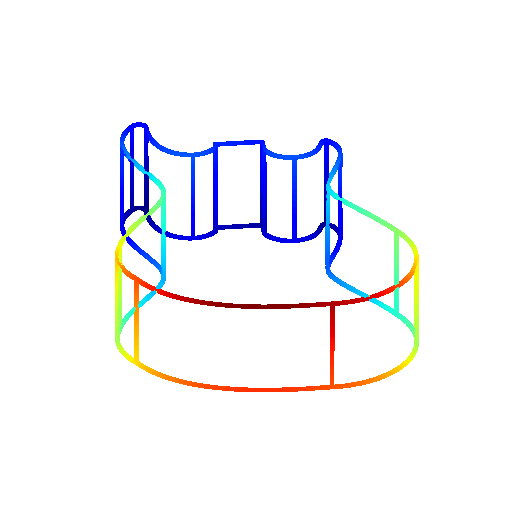}} &
        \raisebox{-.5\height}{\includegraphics[width=0.22\textwidth]{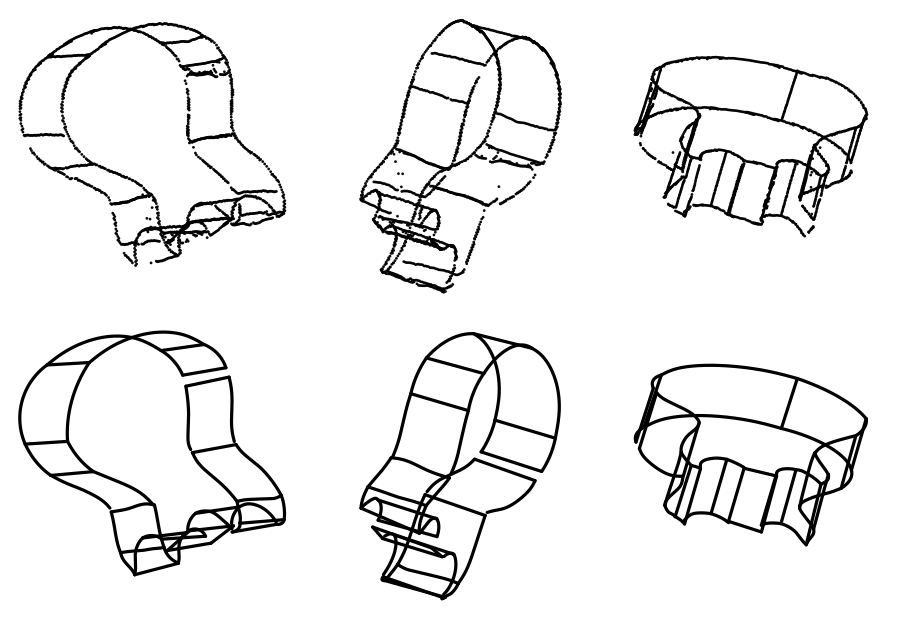}} \\
        \midrule

        \raisebox{-.5\height}{\includegraphics[width=0.07\textwidth]{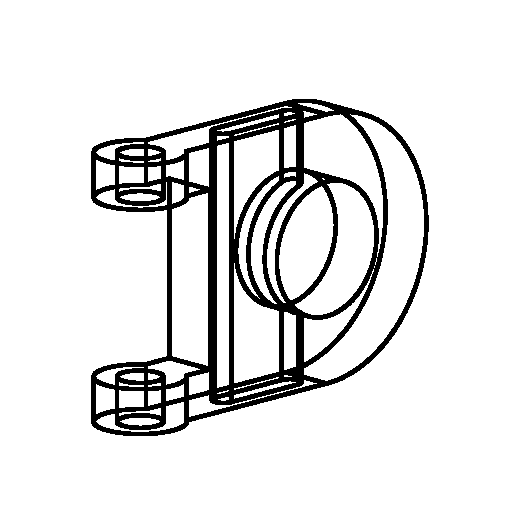}} &
        \raisebox{-.5\height}{\includegraphics[width=0.07\textwidth]{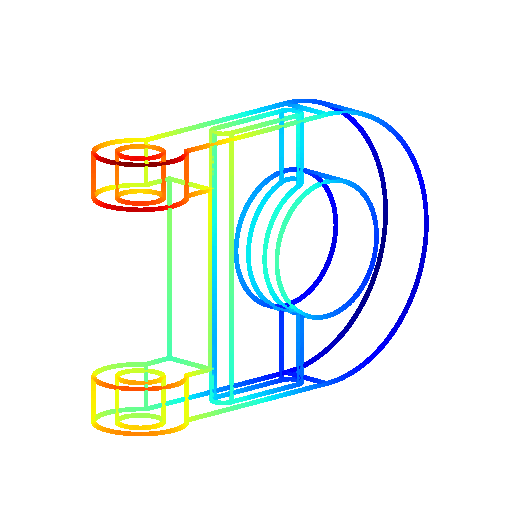}} &
        \raisebox{-.5\height}{\includegraphics[width=0.22\textwidth]{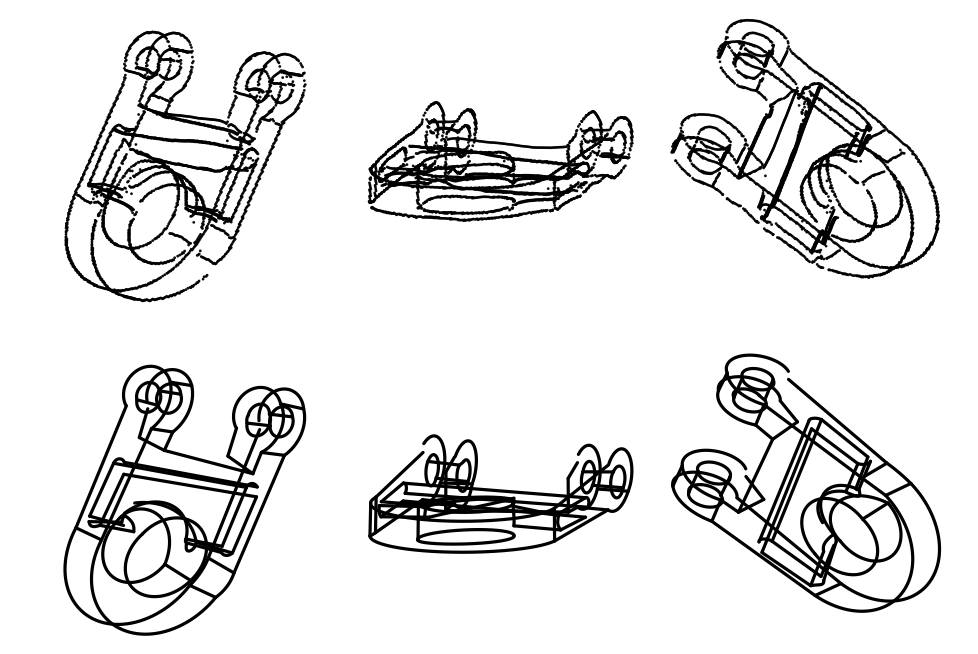}} &
        \raisebox{-.5\height}{\includegraphics[width=0.07\textwidth]{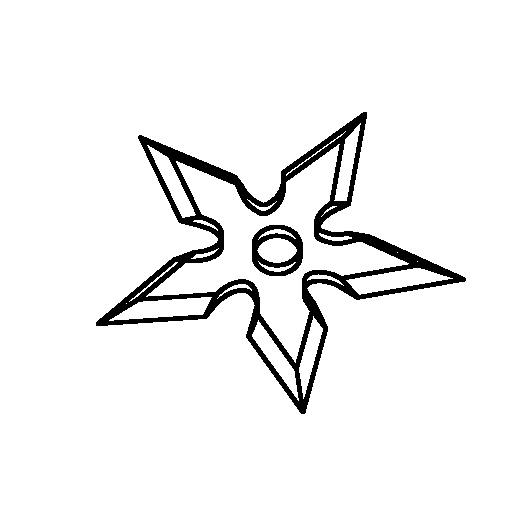}} &
        \raisebox{-.5\height}{\includegraphics[width=0.07\textwidth]{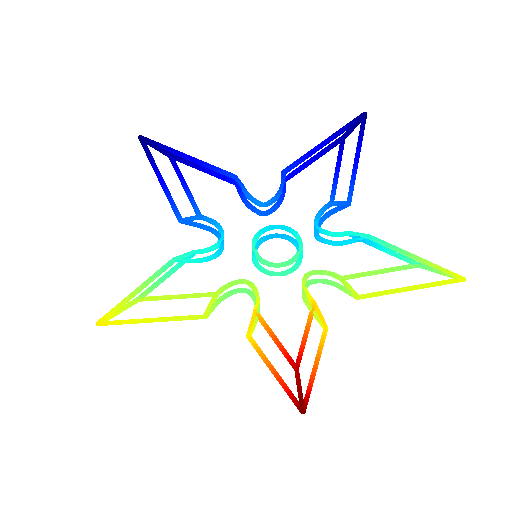}} &
        \raisebox{-.5\height}{\includegraphics[width=0.22\textwidth]{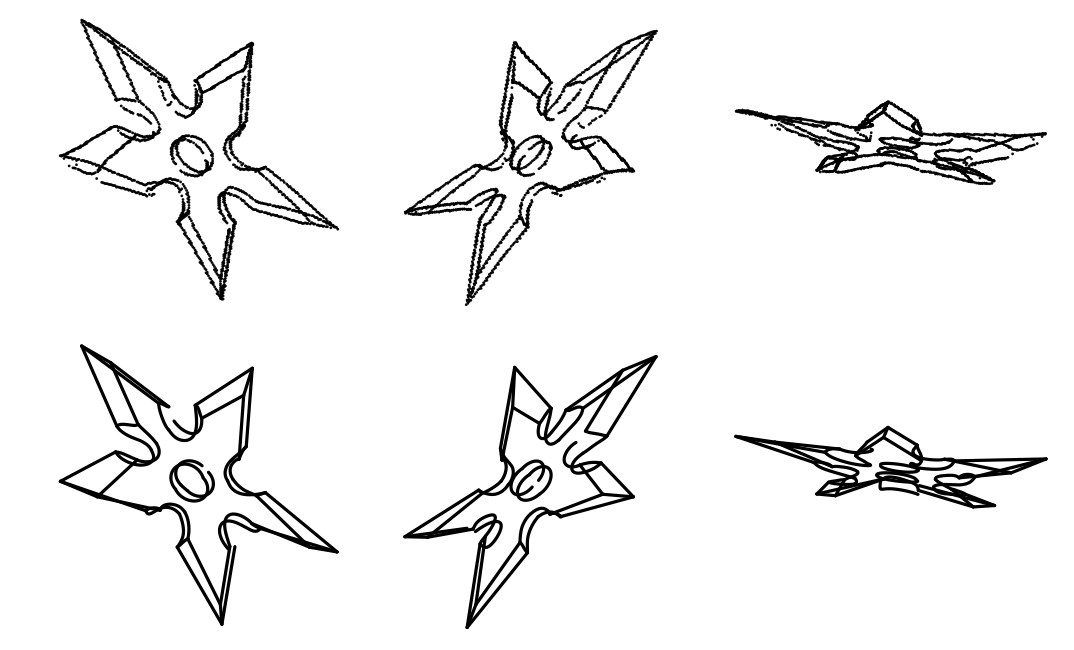}} \\
        \midrule

        \raisebox{-.5\height}{\includegraphics[width=0.07\textwidth]{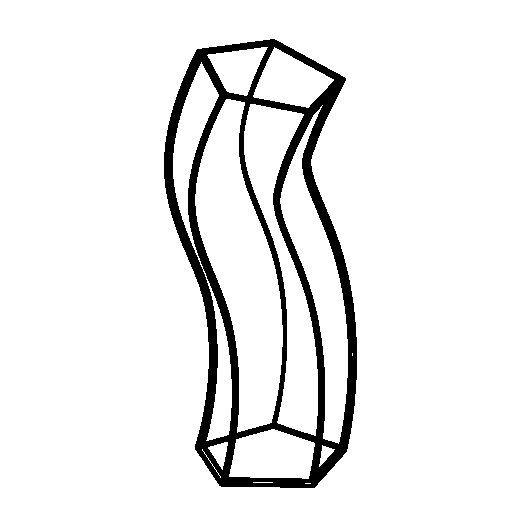}} &
        \raisebox{-.5\height}{\includegraphics[width=0.07\textwidth]{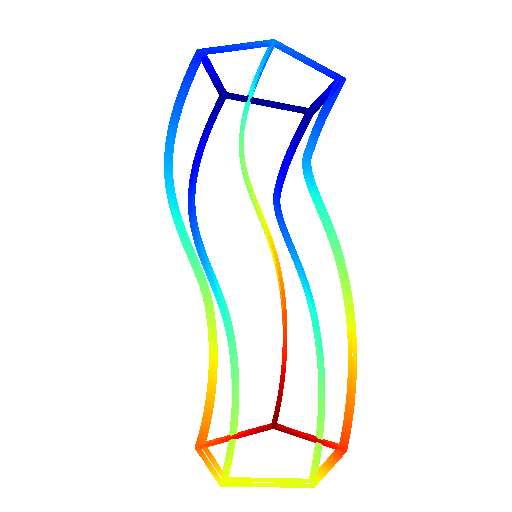}} &
        \raisebox{-.5\height}{\includegraphics[width=0.22\textwidth]{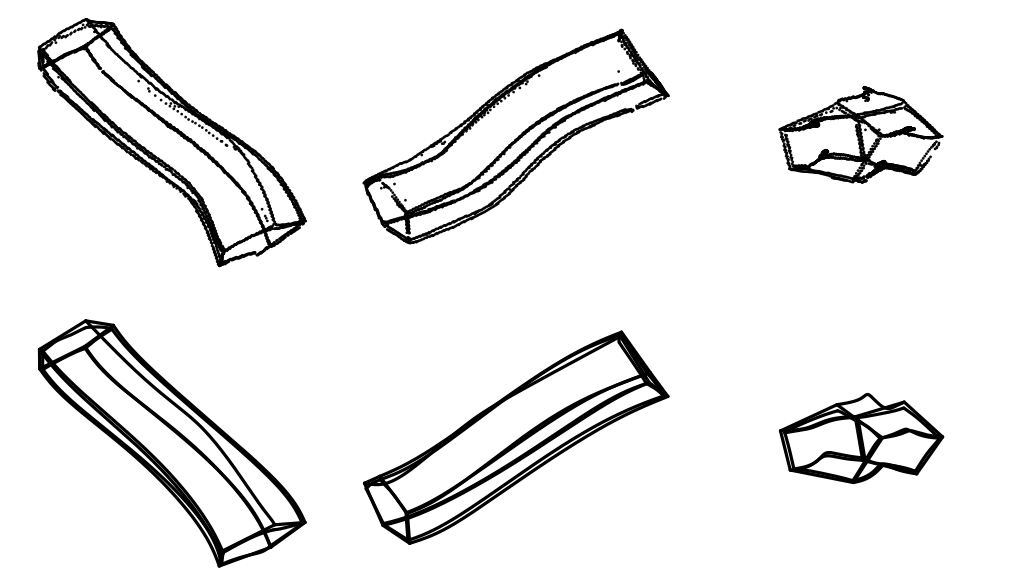}} &
        \raisebox{-.5\height}{\includegraphics[width=0.07\textwidth]{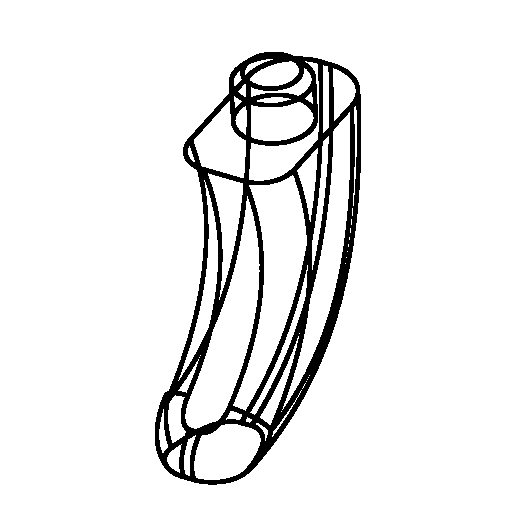}} &
        \raisebox{-.5\height}{\includegraphics[width=0.07\textwidth]{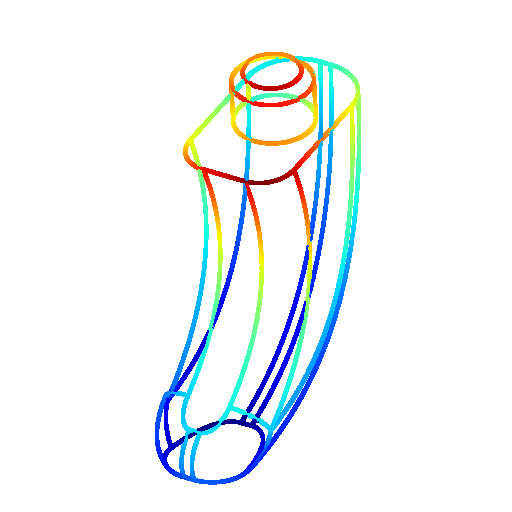}} &
        \raisebox{-.5\height}{\includegraphics[width=0.22\textwidth]{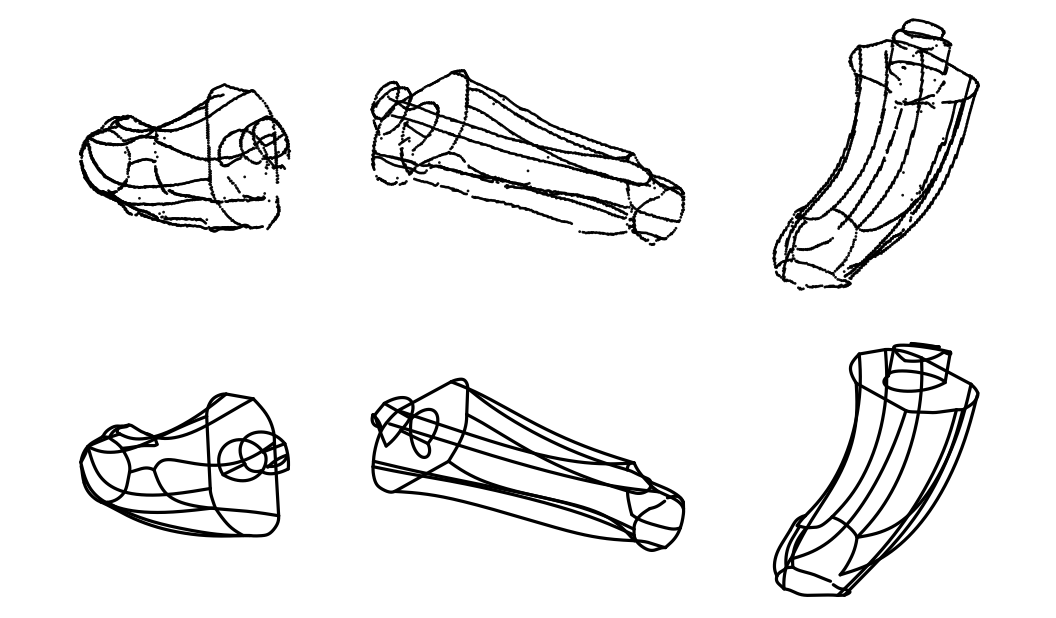}} \\
        \midrule

        \raisebox{-.5\height}{\includegraphics[width=0.07\textwidth]{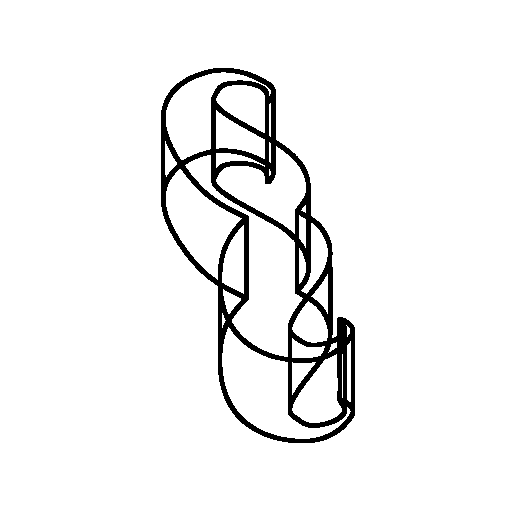}} &
        \raisebox{-.5\height}{\includegraphics[width=0.07\textwidth]{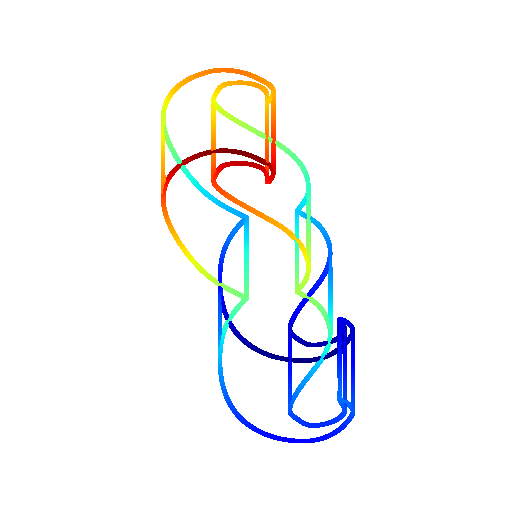}} &
        \raisebox{-.5\height}{\includegraphics[width=0.22\textwidth]{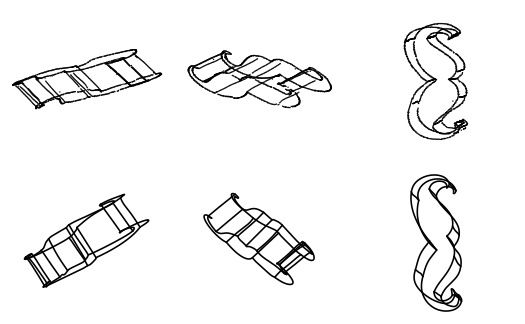}} &
        \raisebox{-.5\height}{\includegraphics[width=0.07\textwidth]{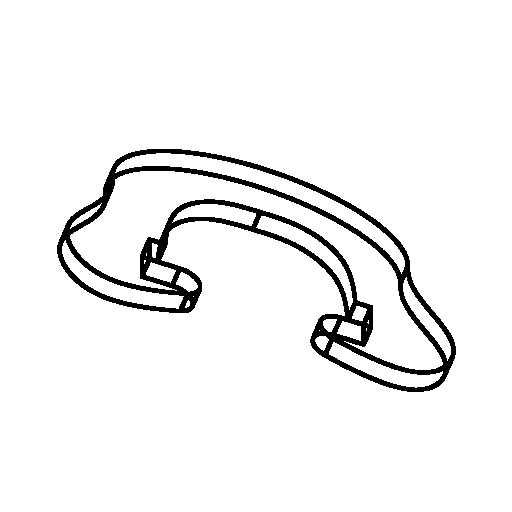}} &
        \raisebox{-.5\height}{\includegraphics[width=0.07\textwidth]{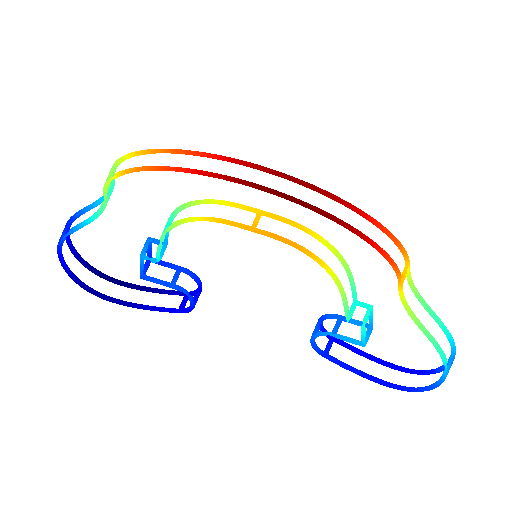}} &
        \raisebox{-.5\height}{\includegraphics[width=0.22\textwidth]{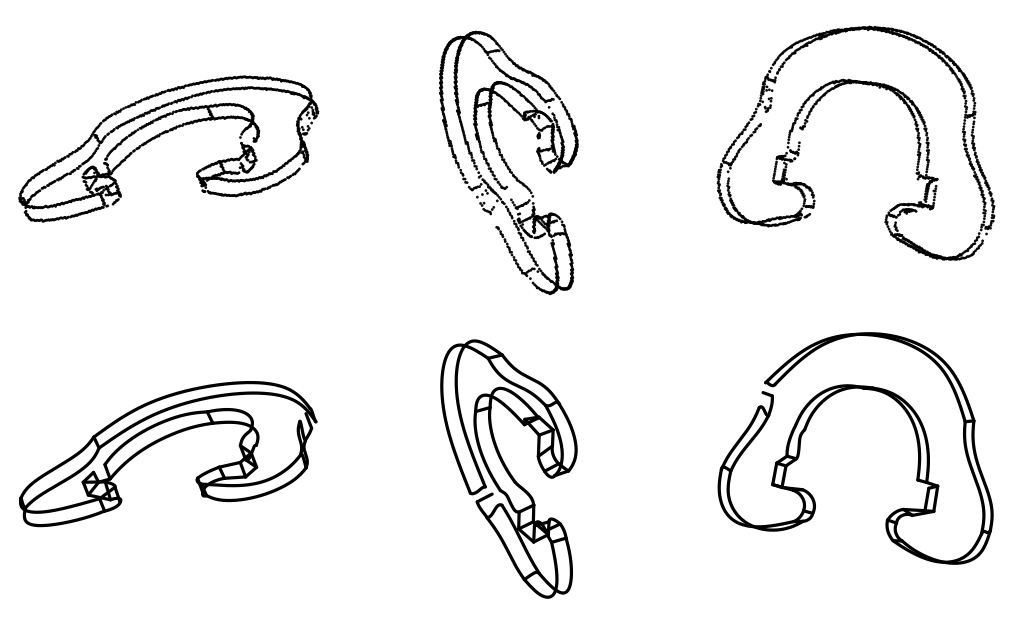}} \\
        \midrule

        \raisebox{-.5\height}{\includegraphics[width=0.07\textwidth]{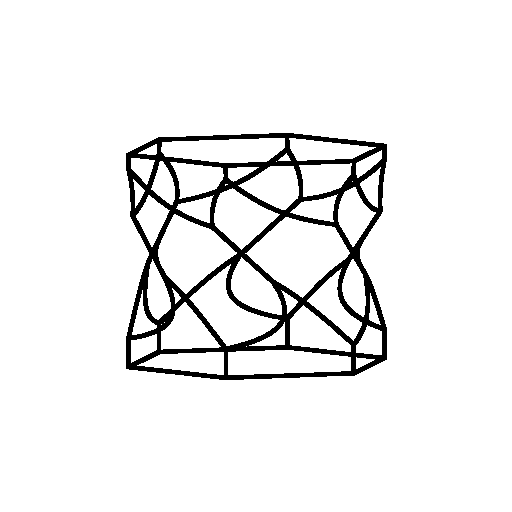}} &
        \raisebox{-.5\height}{\includegraphics[width=0.07\textwidth]{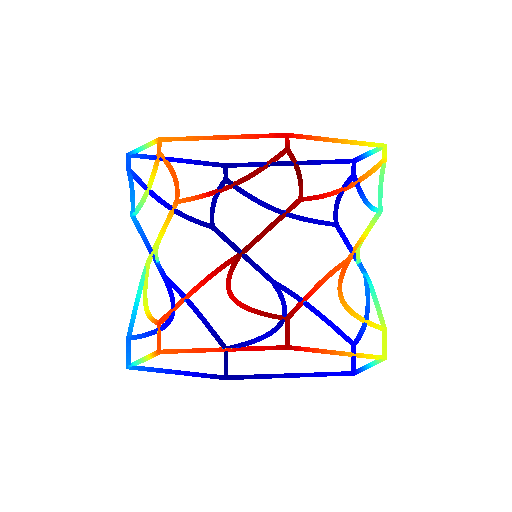}} &
        \raisebox{-.5\height}{\includegraphics[width=0.22\textwidth]{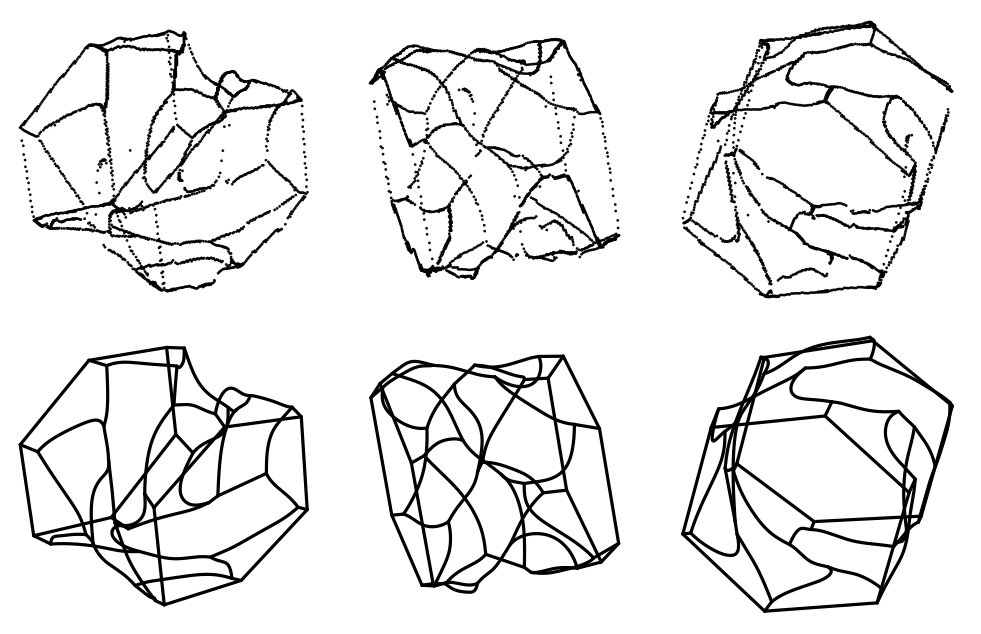}} &
        \raisebox{-.5\height}{\includegraphics[width=0.07\textwidth]{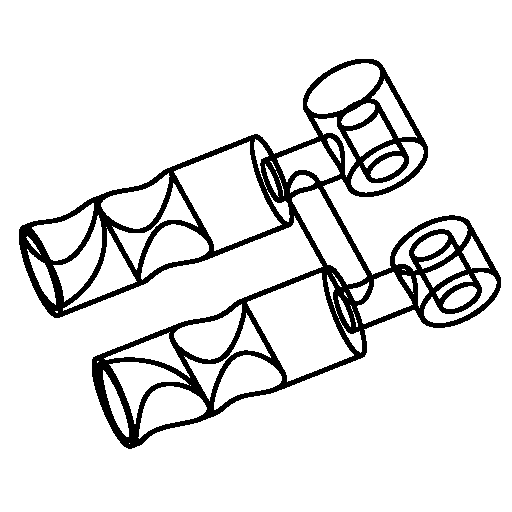}} &
        \raisebox{-.5\height}{\includegraphics[width=0.07\textwidth]{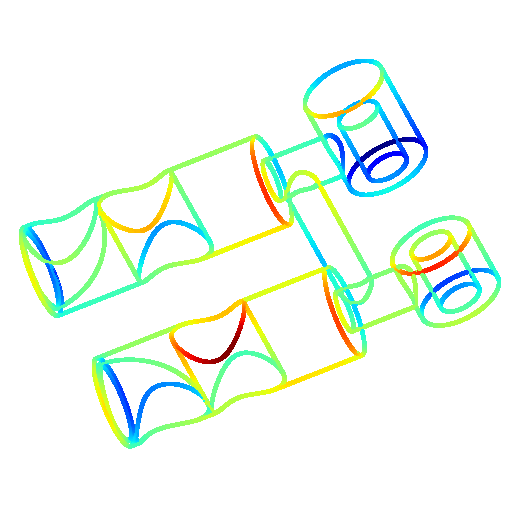}} &
        \raisebox{-.5\height}{\includegraphics[width=0.22\textwidth]{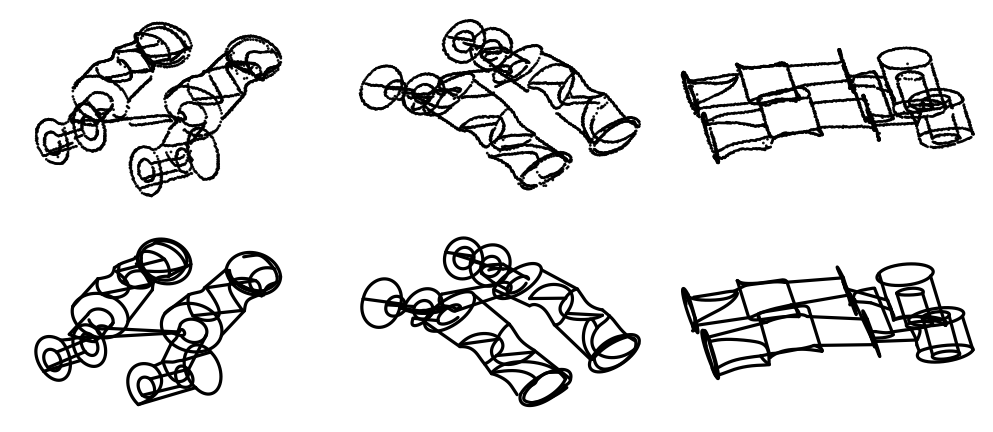}} \\
        \bottomrule
    \end{tabular}

    \caption{\textbf{Qualitative results of our $512 \times 512$ model.} Each block shows the input sketch, the predicted disparity over the drawn strokes, and the 3D reconstruction rendered from three fixed viewpoints at a single shared scale: the raw point cloud unprojected from the predicted depth (top) and the vectorized wireframe fitted with parametric line, circle, and B-spline primitives (bottom). The model reconstructs curved, organic, and thin geometry faithfully; residual errors concentrate in regions dense with overlapping strokes, as in the bottom-right example.}
    \label{fig:qualitative}
\end{figure*}

\subsection{Qualitative Results}
We also evaluate qualitative reconstruction behavior by visualizing predicted depth maps, reconstructed point clouds, and vectorized wireframes (Figure~\ref{fig:qualitative}). The wireframe is constructed by assigning depth-lifted pixels to individual strokes, filtering pixels whose depth is contaminated by occluding strokes, resolving a shared depth value for each junction of the stroke graph, and fitting a parametric primitive (line, circle/arc, or B-spline) to each stroke's surviving 3D support with endpoints pinned to the resolved junctions.

Overall, the model produces high-quality reconstructions across a range of engineering wireframes, including heavily curved and organic geometry. As deduced in the previous sections, residual errors remain concentrated where accidental pixels are dense: the binary input sketch mask cannot differentiate between overlapping 2D strokes, so crossing-heavy regions correspond to the noisiest portions of the reconstructed point clouds (bottom-right example in Figure~\ref{fig:qualitative}). The migration from $256 \times 256$ to $512 \times 512$ conditioning with thinner strokes substantially reduced this failure mode, confirming the resolution hypothesis raised in the v1 analyses of this work, but does not eliminate it entirely for the densest structures.

\section{Limitations and Future Work}
Although our framework presents a highly flexible alternative for the 2D-to-3D reconstruction of wireframes, it also highlights specific challenges inherent to depth-based representations.

\textbf{Sketch input noise.} Human freehand sketches are inherently noisy and highly dependent on an individual's drawing skill. Because our current model is trained exclusively on clean, rendered wireframe projections, it can struggle to generalize to highly irregular or inaccurate freehand input, manifesting in wiggly lines and geometry distortions. To mitigate this, our current interface provides drawing assisting features, such as stroke normalization, endpoint snapping, and an isometric grid, to guide user measurements. Our current formulation preserves the input stroke support by construction, including any user-introduced noise. To truly develop an end-to-end framework capable of capturing a user's underlying design \textit{intent}, future iterations of the diffusion model must learn to map noisy inputs to idealized clean geometries. Training such a model requires perturbing clean wireframes with simulated noise (e.g., vertex jitter and edge wobble). Preliminary experiments suggest a complex spectrum of ``noise,'' ranging from high-frequency stroke perturbations (wobbly lines) to low-frequency structural distortions (e.g., skewing a rectangular face into a parallelogram). Consequently, future models may require explicit noise-conditioning scalars and must learn stronger geometric priors to regularize internal topologies toward what makes a geometrically regular engineering wireframe.

\textbf{Model size and resolution.} Upon scaling our architecture to a larger parameter count, we observed a distinct improvement in our accuracy metrics, and the migration from $256 \times 256$ to $512 \times 512$ (Section~\ref{sec:resolution_scaling}) confirmed that resolution and latent-space fidelity were the binding bottlenecks, roughly halving best-case error. This indicates that our framework benefits from the scaling laws commonly observed in large foundation models, offering a straightforward path to further improvement via increased computational capacity~\cite{bi2024}.

\textbf{Post-processing and mesh conversion.} Our vectorization stage fits parametric primitives (lines, circular arcs, and true B-splines with endpoint-constrained least-squares control points) to the depth-lifted strokes, with occlusion-aware pixel attribution and shared junction depths. Its main remaining weakness is inherited from the depth stage: where overlapping strokes contaminate depth, primitive fits degrade gracefully but visibly. Furthermore, once a mathematically precise wireframe is established, inferring which planar cycles constitute solid faces is required to generate a final, watertight mesh. In this regard, our generative depth approach could be highly complementary when paired with topological sequence models like Neural Face Identification~\cite{wang2022}, or even more traditional face identification techniques like in Shpitalni \& Lipson~\cite{shpitalni1996}. However, aggressive post-processing introduces a representation tradeoff: enforcing rigid parametric constraints diminishes the primary advantage of our generative approach—its ability to model partial, open, and non-standard wireframe structures without predefined vocabularies. Navigating this balance will ultimately depend on the specific downstream needs of the end user.

\textbf{Conditioning input.} To maintain a fully unconstrained pipeline, we deliberately limited our geometric conditioning inputs to binary input sketch masks and partial depth. However, providing explicit stroke-level instance segmentation (i.e., mapping specific pixels to individual strokes and providing the order of strokes drawn) could resolve significant geometric ambiguities. This would alleviate granularity bottlenecks, allowing the network to easily disentangle dense pixel clusters into distinct edges. While requiring users to draw each edge with a single, continuous stroke aligns well with natural human drawing behavior, it reintroduces a subtle operational constraint, lightly compromising the vision of a purely unconstrained predictive pipeline.

\section{Conclusions}
In this paper, we presented a projection-faithful approach to reconstructing 3D geometry from engineering line drawings. Rather than treating sketch interpretation as semantic object completion or CAD program induction, we formulated the task as conditional depth estimation on visible sketch support, allowing a sparse 2D drawing to be lifted directly into an embedded 3D wireframe. This narrower formulation matches an important early-stage engineering need: preserving what the user drew while inferring only the missing depth needed for 3D reasoning. 

Methodologically, the paper shows that latent diffusion with spatial sketch conditioning is a strong fit for this problem. It provides a generative prior for an inherently multimodal reconstruction task, benefits from large-scale rendered engineering data, and naturally supports partial-depth conditioning for iterative design. Empirically, the method becomes substantially more accurate once even limited partial depth is available, validating the sketch-reconstruct-sketch workflow, and scales predictably: expanding to a curve-rich 90,000-shape corpus, retraining the latent autoencoder, and doubling resolution to $512 \times 512$ roughly halves reconstruction error, reaching 3.9\% best-of-five normalized depth error on 4,528 unseen shapes.

Ultimately, this work provides a scalable alternative to traditional parametric modeling by enabling the reconstruction of partial, open, and non-standard wireframe structures without the ``vocabulary bottleneck'' of standard CAD primitives. By reducing the technical barrier between a 2D stroke and its 3D realization, this work moves toward a more intuitive interface where users can fluidly and iteratively ``draw in 3D.''

\section{Acknowledgements}

    This work was supported by the US National Science Foundation (NSF) AI Institute for Dynamical Systems (\href{https://DynamicsAI.org}{DynamicsAI.org}) under grant 2112085. Code is available at \textcolor{VioletRed} {\url{https://eltonc01.github.io/sketch-depth-diffusion}}

{\small
\bibliographystyle{ieeenat_fullname}
\bibliography{references}
}

\end{document}